\documentclass[pdflatex]{sn-jnl}

\usepackage[square,numbers]{natbib}
\jyear{2023}%

\theoremstyle{thmstyleone}%
\theoremstyle{thmstyletwo}%

\theoremstyle{thmstylethree}%

\usepackage{amsmath,amsthm,amssymb}

\usepackage{booktabs, multirow} 
\usepackage{rotating, graphicx}

\usepackage{enumitem}

\usepackage{graphicx}
\usepackage{caption}
\usepackage{adjustbox}

\begin{document}

\title[Human-to-Human Mutual Interaction Recognition]{A Prospective Approach for Human-to-Human Interaction Recognition from Wi-Fi Channel Data using Attention Bidirectional Gated Recurrent Neural Network with GUI Application Implementation}


\author*[1]{\fnm{Md. Mohi Uddin} \sur{Khan}}\email{mohiuddin63@iut-dhaka.edu}

\author[2]{\fnm{Abdullah Bin} \sur{Shams}}\email{ab.shams@utoronto.ca}

\author[3]{\fnm{Md. Mohsin Sarker} \sur{Raihan}}\email{raihan1815505@stud.kuet.ac.bd}

\affil*[1]{\orgdiv{Department of Electrical and Electronic Engineering}, \orgname{Islamic University of Technology}, \orgaddress{\street{Boardbazar}, \city{Gazipur}, \postcode{1704}, \country{Bangladesh}}}

\affil[2]{\orgdiv{The Edward S. Rogers Sr. Department of Electrical and Computer Engineering}, \orgname{University of Toronto}, \orgaddress{\street{10 King’s College Road}, \city{Toronto}, \postcode{M5S 3G4}, \state{Ontario}, \country{Canada}}}

\affil[3]{\orgdiv{Department of Biomedical Engineering}, \orgname{Khulna University of Engineering and Technology}, \orgaddress{\city{Khulna}, \postcode{9203}, \country{Bangladesh}}}

\abstract{
Human Activity Recognition (HAR) research has gained significant momentum due to recent technological advancements, artificial intelligence algorithms, the need for smart cities, and socioeconomic transformation. However, existing computer vision and sensor-based HAR solutions have limitations such as privacy issues, memory and power consumption, and discomfort in wearing sensors for which researchers are observing a paradigm shift in HAR research. In response, WiFi-based HAR is gaining popularity due to the availability of more coarse-grained Channel State Information. However, existing WiFi-based HAR approaches are limited to classifying independent and non-concurrent human activities performed within equal time duration. Recent research commonly utilizes a Single Input Multiple Output communication link with a WiFi signal of 5 GHz channel frequency, using two WiFi routers or two Intel 5300 NICs as transmitter-receiver. Our study, on the other hand, utilizes a Multiple Input Multiple Output radio link between a WiFi router and an Intel 5300 NIC, with the time-series Wi-Fi channel state information based on 2.4 GHz channel frequency for mutual human-to-human concurrent interaction recognition. The proposed Self-Attention guided Bidirectional Gated Recurrent Neural Network (Attention-BiGRU) deep learning model can classify 13 mutual interactions with a maximum benchmark accuracy of 94\% for a single subject-pair. This has been expanded for ten subject pairs, which secured a benchmark accuracy of 88\% with improved classification around the interaction-transition region. An executable graphical user interface (GUI) software has also been developed in this study using the PyQt5 python module to classify, save, and display the overall mutual concurrent human interactions performed within a given time duration. Finally, this article concludes with a discussion on the possible solutions to the observed limitations and identifies areas for further research. Such a Wi-Fi channel perturbation pattern analysis is believed to be an efficient, economical, and privacy-friendly approach to be potentially utilized in mutual human interaction recognition for indoor activity monitoring, surveillance system, smart health monitoring systems, and independent assisted living. 
}

\keywords{
Mutual Human Activity Recognition, Wi-Fi based HAR, Remote Monitoring, Smart Home, Attention BiGRU, Deep Learning
}



\maketitle

\section{Introduction}
\label{sec:Introduction}

Automated Human Activity Recognition (HAR) has gained significant attention due to its versatility in various domains, including automated surveillance and security systems, healthcare monitoring, smart homes, old homes, baby care centers, assisted living for impaired individuals, dormitories, military out-of-bounds areas, criminal detection and digital-forensic investigation, prison cells, and daycare centers for children with special needs such as autism and related disorders. With manual surveillance by a human operator being expensive and sometimes unfeasible, researchers have proposed a multitude of Artificial Intelligence (AI) systems based on Commercial Off the Shelf Wireless Local Area Network (COTS-WLAN), Computer Vision, Wearable Sensor, Radar, and Bluetooth technologies to automate indoor and outdoor human gesture and activity detection. While many of these systems have already been proposed, numerous emerging research projects are still in progress to develop ways to provide point-of-care services that will ensure safety, security, and well-being.

Several studies, including those by Gowda et al. \cite{comboreview7}, Saleem et al. \cite{comboreview6}, Kulsoom et al. \cite{comboreview5}, Gupta et al. \cite{comboreview4gupta}, Islam et al. \cite{comboreview4Islam}, Beddiar et al. \cite{comboreview3}, Zhang et al. \cite{comboreview2}, and Ranasinghe et al. \cite{comboreview1} have conducted extensive surveys of research articles on human activity recognition (HAR). These studies reviewed popular datasets that include spatiotemporal data collected by various sensors, such as accelerometer, gyroscope, RFID, fiberoptic, wearables, WLAN, etc. Additionally, they investigated computer vision and multimodal audio-visual-sensor based recognition systems that deliver real-time notifications to the corresponding supervisor or caregiver situated nearby. The authors reviewed the adopted feature extraction techniques and Artificial Intelligence (AI) driven classification models along with their loss function design and optimization strategies. They compared different approaches to HAR and discussed automated HAR application domains, including commercially funded projects on active-assisted living for the elderly and disabled in smart homes, patient monitoring in medical environments, airport-subway-metrorail surveillance, sport and outdoor, tele-immersion, and more.

In another study, Golestani et al. \cite{MIsensor} developed a wearable magnetic induction (MI) transmitter-receiver coil for human activity recognition. The MI coil is equipped with an L-reversed impedance matching network that uses magnetic induction signals for AI-dependent activity recognition. The study required wearing one receiver MI coil in the subject’s waist and eight transmitter MI coils in the subject’s skeletal bones. Long-Short Term Memory (LSTM) outperformed MI signal variation classification compared to the other five machine learning algorithms used in the study.

However, computer vision incorporates imaging techniques using a video camera that might not be suitable in all circumstances for personal privacy \cite{comboreview4gupta, camprivacy1, camprivacy2, contrst_csi} reasons, especially in indoor residential environments. Moreover, it demands high power, necessary infrastructure, data traffic, and memory usage that might not be economical \cite{comboreview4gupta}, perhaps wastage in many situations. Wearable sensors for human activity detection oftentimes fall short of ensuring proper security \cite{sensorprivacy} since they might expose GPS location or personal health data. In addition, it’s not possible to make obliged everyone to wear such gadgets all the time due to discomfort, health risks, unwillingness, and battery longevity. Radar and ultrasonic technology for human activity detection in the indoor environment require additional electronic devices \cite{radar1, radar2, ultrasonic} to be equipped with, which are consistently not being used in regular standard of living. 

In such context, radio frequency inspection from a WLAN network such as Wi-Fi channel perturbation analysis offers a better economical choice due to its low cost \cite{contrst_csi, csibetter, wifi_attention_better} for human activity recognition since Wi-Fi communication is extensively used nowadays in daily life which is already in the palm of our hand. A Wi-Fi communication device transmits radio signals as electromagnetic waves modulated on an integer number of frequency division multiplexed carriers which are orthogonal to each other. Intuitively, human activities and mutual interactions performed within the line-of-sight Wi-Fi coverage area create distinct changes to the transmitted signal due to the unique pattern of signal reflections from the human body that is performing unique activities. According to Yang et al. \cite{wifi_attention_better}, the use of such WiFi channel perturbations in WiFi-based HAR datasets results in a high level of detail or granularity for better classification compared to datasets based on infrared, radio frequency, RADAR, and environment sensors. Those human activities or mutual interactions can be detected and classified through analysis of the correlation between the signal changes and body movement using signal attributes of a Wi-Fi transceiver pair such as packet delay, coarse-grained Received Signal Strength Indicator (RSSI), accumulated noise, automatic gain control parameter of the receiver circuit, and fine-grained Channel State Information (CSI). The unique distinction among the parametric pattern caused by various human activities can be better detected via artificial intelligence techniques. The overall procedure is neither intrusive nor obtrusive because it does not capture visual data and the entire process runs contactless for safeguarding privacy \cite{wifi_attention_better}. 

Despite plausible advances in the HAR domain, all the existing studies using WLAN RSSI-CSI data are focused solely on recognizing \textbf{independent and single} activities performed by single or multiple subjects. But the recognition of \textbf{reciprocal and concurrent} gestures or interactions between two individuals using WLAN RSSI-CSI data still remains obscure. In the discussion of open issues and challenges related to human activity recognition (HAR), two survey articles by Islam et al. \cite{comboreview4Islam} and Liu et al. \cite{concurrent_composite} have also highlighted the current lack of datasets that include composite and concurrent human activities within a single time-series data sample. Islam et al. \cite{comboreview4Islam} have also reported that most HAR datasets only involve a single activity performed by a single individual, and there is a dearth of datasets that encompass \textbf{mutual single activity performed by multiple persons}, \textbf{mutual concurrent activities performed by multiple persons}, and datasets that incorporate \textbf{inter-activity variability} (i.e., different individuals performing the same activity in slightly different ways). These gaps in the availability of datasets for HAR research pose a challenge to the development of robust and generalizable HAR models that can accurately recognize and differentiate between complex human activities in real-world scenarios. A few computer-vision \cite{videoh2hi1, videoh2hi2, videoh2hi3} and ultra-wideband-radar \cite{radarh2hi} based frameworks for recognizing two-human interactions have been proposed, however.

Cases of physical abuse by babysitters against toddlers or teenagers \cite{babysit}, maltreatment with teenagers in residential care along with forced criminalization \cite{carecenter, youthviolence}, parental control over children left at home by employed parents, incidents of hazing at colleges and dorms \cite{hazing1, hazing2, hazing3}, monitoring at cleanroom of manufacturing lab \cite{cleanroom1, cleanroom2}, monitoring of suspicious and aggressive behavior in a constrained area like prison or restricted zone for service members necessitate continuous surveillance of human-to-human mutual physical interactions while securing the personal privacy, reducing data traffic, and minimizing monetary cost. 

\vspace{2mm}
\noindent Considering the facts mentioned above, 

\vspace{2mm}

\begin{itemize}[noitemsep, topsep=0pt, leftmargin=*]
	\item[i.] To the best of the authors' knowledge, this study represents the \textbf{first-ever} investigation of deep-learning-based recognition of \textbf{human-to-human mutual interaction}, utilizing the \textbf{first-ever publicly available and relevant Wi-Fi dataset} \cite{dataset}. The dataset is distinguished by its inclusion of \textbf{mutuality}, \textbf{concurrency} (i.e., two concurrent mutual activities per time-series sample) as well as \textbf{inter-activity variability}, as discussed in Section-\ref{sec:data_desc}. 

	\item[ii.] It should be noted that the Bidirectional Gated Recurrent Neural Network Unit (BiGRU) deep learning algorithm, when combined with the Self-Attention mechanism, has gained considerable attention for time-sequence data classification and such a combination has certain strengths as mentioned in section-\ref{sec:methodology}. Therefore, the proposed experiment sought to investigate the potential of the novel \textbf{Attention-BiGRU} deep learning model for \textbf{human-to-human mutual interaction} classification in an indoor setting using \textbf{Wi-Fi data}. 


	\item[iii.] First, the investigation was conducted on a single subject pair to develop a greater understanding of the influence of channel metrics diversity of the dataset features. Our proposed deep-learning model demonstrated high-performing mutual interaction recognition results on a single human-subject pair. 

	\item[iv.] The study was further conducted on ten subject pairs for validating the high performance under influence of variegated subject pairs and clearly, this also exhibited reasonable classification performance indicating that the model can be trained and applied for upto ten subject pairs. 

	\item[v.] But the cross-test experiment on new subject pairs did not produce satisfactory results which indicates that profiling of mutual interaction features from untrained subject pairs protrudes more complicated channel metrics diversity. 

	\item[vi.] Alongside the proposed solution, further discussion on such an undesirable result during the cross-test experiment pointed out the possible pitfalls and provided suggestions that need to be taken care of in further research. 

	\item[vii.] To develop an overall software framework and visually depict the classifications, the study finally developed a graphical user interface (GUI) executable (not standalone) software designed using PyQt5 \cite{pyqt5} GUI app builder Python module.
\end{itemize}

\vspace{2mm}
Remaining of the article is organized as follows: A comprehensive review of the existing literature on WLAN based Human Activity Recognition (HAR) is presented in Section-\ref{sec:Literature_Review}. Section-\ref{sec:data_desc} delivers an overview of the dataset and challenges with the recorded pattern of the time-series data values for different mutual interactions. Section-\ref{sec:wlan_theory} describes the feature variables provided with the dataset. Experimental framework, proposed deep-learning model description and executable GUI software description are introduced in section-\ref{sec:methodology}. Research outcomes and perceived hindrances are discussed in section-\ref{sec:result}. Future research prospects are narrated in section-\ref{sec:scopes}. The article is concluded with section-\ref{sec:conclusion} providing the supplementary materials in section-\ref{sec:supp_link}.

\section{Literature Review}
\label{sec:Literature_Review}
For a comprehensive understanding of Human Activity Recognition (HAR) from WLAN data, it is essential to have an in-depth knowledge of the existing literature on such a topic. Therefore, this section aims to provide a detailed overview of the research that has been done so far in this area and identify the gaps that need to be addressed. Through this review, this article aims to contribute to the existing body of knowledge on HAR from WLAN data and provide a foundation for the objectives and methodology of this study.

A survey report \cite{csi_review} discussed the pros and cons of different Internet-of-Things (IoT) protocols with research advancements in RF-based gesture or activity recognition. Authors compared the performance of recent state-of-the-art proposals on body-sensor, video camera, WLAN Received Signal Strength Indicator (RSSI), and Channel State Information (CSI) based single or multiple human independent activity recognition models incorporating various signal processing dependent pre-processing and feature extraction techniques with their application and performance. Authors of the survey report confined their maximum focus on the WLAN approaches due to its immense potential \cite{CSI5G} and narrated that the recognition accuracy with CSI is way higher than RSSI of Wi-Fi signal because CSI provides amplitude and phase information for each subcarrier frequency of OFDM modulation per packet of the multiple-input multiple-output (MIMO) transmission. But RSSI provides only a single magnitude value per packet. The article further shed light on both the processed and raw CSI data analysis using mathematical model-based approaches \cite{modelbasedcsi_review} like CSI speed-activity-mobility-ratio-quotient model, Angle of Arrival model, Fresnel Zone model along with learning-based approaches dependent on feature extraction, machine and deep learning. Deep learning approaches, especially Convolutional Neural Network (CNN), have become much popular due to their automated ability of feature extraction from raw CSI data with a large volume of data handling capacity even though it is observed previously to have more publications with machine learning. Recurrent Neural Network (RNN) models like vanilla RNN and LSTM are also being adopted \cite{ieeelstm1, ieeelstm2} nowadays by few researchers because of their time-sequence classifying propensity. Based on a comprehensive literature survey, Gupta et al. \cite{comboreview4gupta} have also reported that in recent years (2017-2021), device-free human activity recognition (HAR) research has gained significant attention in utilizing Wi-Fi Channel State Information (CSI) dataset. According to Gupta et al. \cite{comboreview4gupta}, this time period also coincides with the growing popularity of various deep learning techniques in the HAR domain, despite the earlier HAR models being largely reliant on camera-based or customized wearable sensor data.

Damodaran et al. \cite{ubuntucsi} conducted a study on walk-run-sit-stand-empty room detection at the living room and hallway environment using two laptops running on 64 Bit Ubuntu operating system equipped with Intel 5300 NIC served as 1x3 Tx-Rx system. Authors extracted CSI values using Linux 802.11n CSI Tool \cite{csi_tool} and only the CSI amplitude is considered for analysis. They proposed two models for activity recognition, viz. Support Vector Machine (pre-processing done using discrete wavelet transform for noise reduction, principal component analysis for dimensionality reduction, and feature extraction conducted using power spectral density, frequency of center of energy, Haart wavelet analysis), and LSTM (raw data used after denoising via discrete wavelet transform). 

Nonetheless, Ashleibta et al. \cite{CSI5G} pointed out, also reported by Damodaran et al. \cite{ubuntucsi}, a major drawback of Network Interface Cards (NIC) used for controlling the flow of the data over Wi-Fi link and channel data estimation. IEEE 802.11n compliant Wi-Fi access points use 51 subcarriers during OFDM modulation but Intel 5300 NIC at the receiver side can report data for the utmost 30 subcarriers while CSI information of the remaining 21 subcarriers is lost. Simultaneous use of NIC for networking function and CSI estimation using external tools for human activity recognition may hamper the reliability of NIC’s long-term performance as well. To avoid these problems, Ashleibta et al. \cite{CSI5G} claimed to be the pioneering developer of human occupancy (the number of people in the room) counting and their parallel multi-activity (performing at 4 distinct positions of the room) recognition via deep 1D-CNN model using CSI (amplitude only) generated by two Universal Software Radio Peripheral (USRP) \cite{usrp1, usrp2} formed Single-Input Single-Output (SISO) communication system operating at 3.75 GHz carrier frequency which falls under the 5G frequency band (3.4-3.8 GHz). Their generated dataset consists of a total of 1,777 trials having 16 combinations of 3 distinct activities (sitting, standing, walking) along with empty rooms performed by 4 subjects in a lab setting. The developed recognizer model comprises a data pre-processing step formed of mean CSI amplitude finder across all 51 subcarriers, fourth-order Butterworth lowpass filter for noise reduction, 3-level discrete wavelet transform calculator followed by the classifier model formed of 13 1D-CNN layers, and 5 max pooling layers. Authors reported the accuracy of the parallel multi-activity recognition experiment to be around 80\% and the accuracy of the occupancy experiment ranging between 86\%-95\%. They also justified the superiority of their USRP-formed 5G-enabled SISO system because of more discernible CSI pattern formation compared to a WiFi-NIC system operating at a 5 GHz frequency band.

In the study by Shalaby et al. \cite{csiRev_cnnGRUattention}, a Single Input Multiple Output (SIMO) communication link was established between a WiFi router and an Intel 5300 NIC mounted in a laptop to classify six independent non-concurrent human activities  (i.e. lie down, fall, walk, run, sit down, and stand up) using CSI amplitude values from an unbalanced dataset. Four deep learning architectures, namely CNN-GRU, CNN-GRU-Attention, CNN-GRU-CNN, and CNN-LSTM-CNN, were explored, with the CNN-GRU model performing the best and achieving over 99\% in all performance metrics.

Yang et al. \cite{autofi} utilized two TPLink-N750 WiFi routers to capture CSI amplitude data from a 5 GHz channel frequency 40 MHz carrier bandwidth WiFi 802.11n signal in SIMO mode. They employed self-supervised learning followed by few shot learning to classify eight independent non-concurrent human activities (i.e. up-down, left-right, pull-push, clap, fist, circling, throw, and zoom) from the CSI amplitude data, achieving gesture recognition accuracy of 0.8971 and gait analysis accuracy of 0.8333 using 3-shot learning. According to the authors, the use of self-supervised and few shot learning techniques ensures that the recognition model is unbiased towards the data during supervised learning and can adapt to new data.

Bocus et al. \cite{slfspcsi} investigated the Multiple Input Multiple Output (MIMO) radio link from two 3-antenna Intel 5300 NICs to classify six independent non-concurrent human activities (i.e. lay down, sit, stand, stand from the floor, walk, body rotate) from CSI amplitude data. They pre-processed the data through denoising using Discrete Wavelet Transform (DWT), segmentation, and dimensionality reduction using Principle Component Analysis (PCA) techniques before converting them to multiview spectrogram images using Short Time Fourier Transform (STFT). The images were then classified using AlexNet, VGG16, and ResNet, with AlexNet achieving the best results, obtaining a macro F1 score of 0.68.

Xu et al. \cite{contrst_csi} conducted research on various independent non-concurrent human activities from three distinct datasets, namely Office Room having 7 classes, Falldefi having 2 classes, and SignFi having 276 classes. The authors integrated time information with the CSI data and developed a novel mutual information enhanced dual-stream contrastive learning unsupervised framework. The proposed framework achieved impressive results, with 94.6\% accuracy and F1 score on the Office Room dataset, 80\% accuracy and F1 score on the Falldefi dataset, and approximately 87\% accuracy and F1 score on the Signfi dataset.

Al-qaness \cite{tailorcsi} investigated the use of a MIMO scheme with a TP-Link WiFi router and Intel 5300 NIC mounted in a laptop to classify nine independent non-concurrent human activities, namely dodge, push, strike, circle, punch, bowl, drag, pull, and kick. To extract features from the CSI data, the authors utilized both the amplitude and phase angle, which were then subjected to a Butterworth lowpass filter for denoising. Next, Principle Component Analysis was applied to the denoised features followed by pattern segmentation using Hilbert-Huang transform, which determined the width of the dynamic time window. Finally, the authors extracted 12 statistical features from the preprocessed data and trained a Random Forest Classifier, which achieved precision ranging from 90\% to 97\%, recall ranging from 88.35\% to 100\%, and F1-score ranging from 89.67\% to 98.48\%.

Although the field of learning-based human activity recognition (HAR) has made significant strides, recent literature reveals that the majority of studies have focused on using a SIMO communication link between two Intel 5300 NICs installed in laptops or two WiFi routers, with most articles classifying less than ten independent non-concurrent human activities. Therefore, it is essential to advance the research on HAR by incorporating a MIMO communication link between a WiFi router and an Intel 5300 NIC installed in a computer to simulate real-life scenario. Moreover, to improve the performance and robustness of HAR models, it is necessary to develop learning-based models that can classify mutual concurrent human-to-human interactions conducted over varying durations. Additionally, it is important to differentiate the same types of interactions performed by either the left or right hand/leg. In this article, the authors took the initiative to work on these areas to advance the field of HAR research.

Li et al. \cite{radarBiGRU} delineated the superiority of the Bidirectional Gated Recurrent Neural Network Unit (BiGRU) model on single-human independent-activity recognition compared to RNN, BiRNN, and GRU model; though the authors used `wideband radar waveform' obtained from RADAR High-Resolution Range Profile signal. Because the BiGRU deep learning algorithm accompanied by the Self-Attention mechanism has recently gained considerable attraction for time-sequence data classification such as HTTPS traffic classification \cite{grutrend1}, EEG-based emotion classification \cite{grutrend2}, P300 EEG signal classification \cite{grutrend3} and because the model has certain strengths as mentioned in section-\ref{sec:methodology}, this study aimed to investigate the potential of the \textbf{Attention-BiGRU} deep learning model for \textbf{human-to-human mutual interaction} classification in an indoor setting using \textbf{Wi-Fi channel information data}.

\section{Dataset Description}
\label{sec:data_desc}

To explore the philosophy of human-to-human interaction recognition via Wi-Fi data classification using Artificial Intelligence (AI), this study opted to take advantage of \textit{`the only available first-ever published Wi-Fi-based human-to-human interaction dataset'} by Alazrai et al. \cite{dataset} contemplating 13 activities performed by two individuals in the central area between 4.3 meters apart Wi-Fi access point and Network Interface Card (NIC) kept in Line-of-Sight (LOS) manner targeting each other in an indoor environment (5.3 square meters furnished room). A publicly available CSI tool \cite{csi_tool} had been used to record Wi-Fi data and the device configurations are depicted in Figure-\ref{fig: MIMO}.

The dataset comprises 40 folders for interactions performed by 40 different subject-pairs. Each folder contains 12 sub-folders for 12 different interactions. Each sub-folder contains 10 different data (*.mat) files containing Wi-Fi packet data [viz. Wi-Fi packet arrival timestamp, number of transmit and receive antennas (NTx, NRx), channel noise, automatic gain control (AGC in dB), received signal strength indicator (RSSI in dB) at the 3 receive antennas, complex valued channel state information (CSI) and interaction label] for 10 trials of each interaction. The complex valued CSI array is of order (NTx, NRx, NSc) = (2, 3, 30) where, NSc = Number of Subcarriers used during OFDM modulation.

It's to be noted that each trial is composed of 2 interactions (steady-state and any of the 12 interactions), that’s how 13-interactions altogether. The 13 human-to-human interactions are: steady-state, approaching, departing, handshaking, high-five, hugging, kicking (left-leg), kicking (right-leg), pointing (left-hand), pointing (right-hand), punching (left-hand), punching (right-hand), and pushing.

It's also to be noted that, the data files are of unequal time-series length, i.e. time-series data ranging upto 1040-2000+ Wi-Fi packets and packet arrival time-difference are also seen unequal. This occurred due to the following data acquisition time-length: 

\begin{itemize}
\setlength\itemsep{0em}
    \item[--] 2.00 second for steady-state + 3.00 second for hugging or kicking or punching
    \item[--] 3.50 second for approaching + 2.00 second for steady-state (Note that, steady-state performed at the end portion only for approaching action)
    \item[--] 2.00 second for steady-state + 3.50 second for departing
    \item[--] 2.00 second for steady-state + 4.00 second for handshaking or high-five or pushing
    \item[--] 2.00 second for steady-state + 4.50 second for pointing
\end{itemize}

\vspace{6pt}
Such kind of \textbf{time-length inequality}, two \textbf{mutual interactions} merged in each trial (i.e.: \textbf{mutuality and concurrency}), steady-state either at the \textbf{beginning} or at the \textbf{end} portion of the trial, the difference in height-weight-clothing-gesture style of 66 subjects (i.e.: \textbf{inter-activity variability}) added an extra layer of complexity to the \textbf{LOS-faded} Wi-Fi signal pattern-recognition AI challenge.

\section{Description of the Dataset's Feature Variables}
\label{sec:wlan_theory}

\begin{figure*}[ht]
\centering 
\resizebox{\textwidth}{!}{\includegraphics{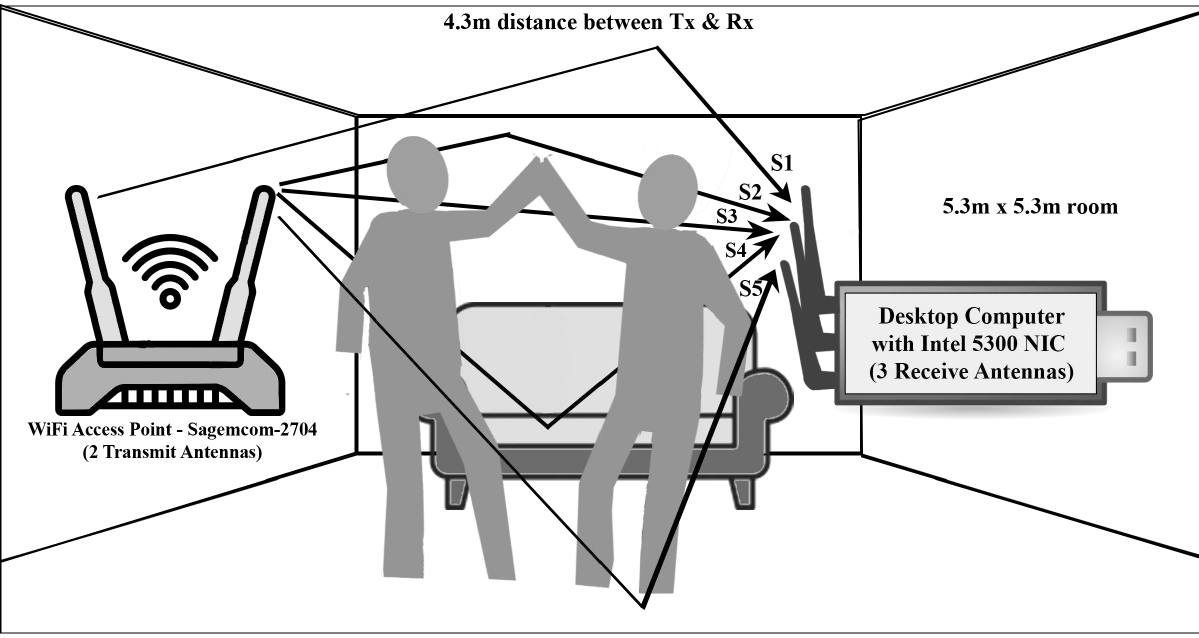}}
\caption{{Conceptualization of the 2x3 MIMO-OFDM indoor WLAN transmission system in case of two-human line-of-sight interaction. WiFi AP specifications are 2.4GHz frequency band, wireless channel no.-06, 20MHz channel bandwidth, modulation coding scheme index-08, MIMO OFDM (30 subcarriers) and IEEE 802.11n compliant device. Here, S3 represents Line-of-Sight (LOS) signal fading due to two human mutual interaction and Non-Line-of-Sight (NLOS) signals are reflected from Ceiling (S1), Wall (S2), Indoor Object (S4), Floor (S5)}}
\label{fig: MIMO}
\end{figure*}

To attain the dataset, an IEEE 802.11n standard \cite{ IEEE80211n} (released in 2009) compliant 2x3 MIMO-OFDM (Appendix-\ref{appendix:mimo_ofdm}) wireless LAN communication system had been used during data collection. Figure-\ref{fig: MIMO} offers an exposition on the system configuration used during data collection \cite{dataset}. 

This section provides a brief overview of the feature variables included to the dataset, the environmental factors that affect and underlie the pattern changes of those feature values when applied to indoor WLAN environments, the mathematical representation of some of those features, and some observations on a few variables in recent WLAN research works. These also illustrate the significance of using these feature variables in the experiments of this research work in order to make use of feature pattern changes to classify 13 distinct mutual human-to-human interactions using the proposed Attention-BiGRU deep learning model.

\subsection{Signal Fading and Noise}
\label{subsec:sig_fad_noise}

Unlike this particular experiment where the Line-of-Sight (LOS) faced transmitter and receiver are stationary at fixed distance (4.5 meter apart), the transmitted electromagnetic (EM) signal encounters attenuation (amplitude deviation, phase change and delay spread from multipath time-delay) due to large-scale fading (i.e. indoor path loss) and small-scale fading (i.e. absorption-reflection-diffraction-scattering-interference during multipath indoor propagation and speed of surrounding objects viz. movement of two human during interaction) across the propagation range of Wi-Fi signal from transmitter to receiver. The wavelength ($\lambda = 125$ mm, eq$^{tn}$-1) of the Wi-Fi signal with $f = 2.4 GHz$ carrier frequency is too small compared to 4.5m Tx-Rx distance such that EM wave propagation phenomena and equations sustain for the indoor environment.
	{\footnotesize \[\tag{1}\lambda =\frac{c}{f}=\frac{3\times {{10}^{8}}m{{s}^{-1}}}{2.4GHz}=\frac{3\times {{10}^{11}}mm{{s}^{-1}}}{2.4\times {{10}^{9}}{{s}^{-1}}}=125mm\text{\;\; ;\; here,\;}c=\text{\;speed\;of\;EM\;wave}\]}

Mean received power logarithmically reduces with distance in case of indoor path loss; hence if path loss is denoted by $P_{Loss}$, path loss exponent by n, transmitter-receiver separation by ${{d}_{Tx\leftrightarrow Rx}}$ and 1m-100m reference distance by $d_{ref}$, then free-space indoor path loss for single room can be equated from log-distance path loss model as,
{\small \[\tag{2}{{P}_{Loss}}\left( dB \right)={{P}_{Loss}}\left( {{d}_{ref}} \right)+10n\log \left( \frac{{{d}_{Tx\leftrightarrow Rx}}}{{{d}_{ref}}} \right)\text{\; ;\; here,\;}n=2\text{\;for\;free\;space}\]}

Transmitted EM wave strength sometimes gets partially absorbed by indoor objects like spongy-foamy couch, soft curtains, water (if any, e.g. aquarium), concrete-wall, brick-wall, metal-wall, electronic appliances (computer, refrigerator, microwave oven etc.). While obstructed by ceiling, floor, wall, humans with different weight-height and flat indoor objects, EM signal gets reflected and a trifling amount of wave bounces towards any other direction than the receiver, thus reducing received signal energy. Both partial absorption and reflection may occur simultaneously in varying amount depending on material composition and texture of the obstacle and incident angle of the signal on the obstacle. 

Multiple Lambertian reflected waves generate after the EM wave gets scattered from uneven rough surface like indoor ornamental plant leaves, curly short human-hairs along direct LOS path, dust or smoke or micro-water-droplets present in the air etc. Secondary wavelets originate maintaining Huygen’s principle while the transmitted EM beam gets diffracted at sharp edges (corner of walls, desk, tables, chair, ceiling fan etc.) and propagate along their split direction curving around that obstacle. These multipath propagated signals exhibit signal amplitude deviation and phase shift while received by the receiving antennas.

Transmitted signal might also encounter interference with any other radio wave (signal from another Wi-Fi router in other room, microwave oven, cellphone etc. if operating at 2.4 GHz carrier frequency band with unique modulation frequency). This also adds Additive White Gaussian Noise (AWGN) to the transmitted signal. \cite{rappaport, daniel, IndoorWLAN}

Since the Tx-Rx has LOS path (eventhough sometimes partially and randomly obstructed due to two human interactions), the small-scale fading can be simplistically expressed in a generalized fashion using Rician Fading model \cite{shankar} considering sinusoidal transmission signal having amplitude A, at a carrier frequency of $f = 2.4 GHz$ as follows,

\vspace{6pt}
\noindent Received Signal,
{\footnotesize \[\tag{3}\begin{array}{l}
  {{R}_{Rician}}(t)={{A}_{LOS}}\cos (\omega t)+\sum\limits_{i=1}^{\#multipaths}{{{A}_{i}}\cos (\omega t+{{\phi }_{i}})\text{\;;}} \\
  
  \;\;\;\;\;\;\;\;\;\;\;\;\;\;\;\;\;\;\;\;\;\;\;\;\;\;\text{\;\;\;\;\;here,\;angular freq, \;}\omega =2\pi f\text{\;and\;}\phi = \text{signal\;phase\; difference} \\
  
 \;\;\;\;\;\;\;\;\;\;\;\;\;={{A}_{LOS}}\cos (\omega t)+\cos (\omega t)\sum\limits_{i=1}^{\#multipaths}{{{A}_{i}}\cos ({{\phi }_{i}})}-\sin (\omega t)\sum\limits_{i=1}^{\#multipaths}{{{A}_{i}}\sin ({{\phi }_{i}})} \\ 
 
 \;\;\;\;\;\;\;\;\;\;\;\;\;=(M+{{A}_{LOS}})\cos (\omega t)-N\sin (\omega t)\text{\;\; ;} \\
 
 \;\;\;\;\;\;\;\;\;\;\;\;\;\;\;\;\;\;\;\;\;\;\;\;\;\;\text{\;\;\;\;\;here,\;}M =\sum\limits_{i=1}^{\#multipaths}{{{A}_{i}}\cos ({{\phi }_{i}})}\text{\;\;and\;\;}N = \sum\limits_{i=1}^{\#multipaths}{{{A}_{i}}\sin ({{\phi }_{i}})}
\end{array}\]}

\noindent Received Power, 
{\small \[\tag{4}P={{(M+{{A}_{LOS}})}^{2}}+{{N}^{2}}\]}

\noindent Average Received Power, 
{\footnotesize \[\tag{5}\bar{P}=2{{\sigma }^{2}}+A_{LOS}^{2}\text{\;\;\; ; \;here,\; variance}:\text{\;}{{\sigma }^{2}}=\frac{\sum\limits_{i=1}^{\#multipaths}{{{({{x}_{i}}-\bar{x})}^{2}}}}{\#multipaths};\text{\;}{{x}_{i}}={{A}_{i}}\cos \left( \omega +{{\phi }_{i}} \right)\]}

\noindent Rician factor, {\small \[\tag{6}K=\frac{A_{LOS}^{2}}{2{{\sigma }^{2}}}\]}

\noindent If B(.) is the modified Bessel function of first kind \cite{besselfn1, besselfn2}, U(.) is the unit step function, the Probability Density Function (PDF) of the Received Power,
{\small \[\tag{7}PD{{F}_{Rician}}(p)=\frac{K+1}{{\bar{P}}}\exp \left( -K\left( K+1 \right)\frac{p}{{\bar{P}}} \right)B\left( 2\sqrt{\frac{K\left( K+1 \right)}{{\bar{P}}}}p \right)U\left( p \right)\]}
	
Andjamba et al. \cite{namibiainterference} published a report on inter-network interference suffered by on-campus IEEE 802.11 standard access point operating at both 2.4 and 5 GHz carrier frequency band after two months of monitoring due to over 345 IEEE 802.11b/g/n/ac access points surrounded off-campus. Liu et al. \cite{InterNetworkInterference} proposed a novel mathematical architecture via theoretical analysis and developed an opensource program for IEEE 802.11 standard one-to-one Wi-Fi network’s throughput ascertainment under the repercussion of inter-network-interference due to Wi-Fi devices of similar kind. Authors verified the analytical expressions along with the developed program using three network structures (string topology with 3 and 4 one-to-one networks, grid topology with 9 one-to-one networks) wherein they used `airtime concept’ for finding out each network throughput which Wan et al. \cite{airtimeconcpet} believe handy for divulging the packet sensing duration at each node and frame collisions due to hidden nodes.

Chandaliya et al. \cite{BTInterference} observed the effect of signal interference on IEEE 802.11n standard Wi-Fi device throughput due to 2-pairs of Bluetooth devices sharing data in the vicinity of Wi-Fi range at right-angled spatial position with respect to Wi-Fi LOS link, all operating at 2.4 GHz carrier frequency band. Practical presence of Bluetooth devices decreased the Wi-Fi throughput by 6.27 dB, increased packet-arrival-time-difference by 0.26 second and increased packet-drop-rate by 2.73 times compared to ideal software simulation of Wi-Fi data sharing alone. Natarajan et al. \cite{multiInterference} also performed mathematical analysis, physical layer analysis and Medium Access Control (MAC) layer analysis on the cross-technology signal interference while in coexistence among IEEE 802.15.4, Bluetooth Low Energy (BLE) and IEEE 802.11 technologies.

\subsection{Packet Arrival Time Difference}
\label{subsec:timediff}
Sarker et al. \cite{peoplemovement} conducted an investigation using IEEE 802.11 technology at multitude of environments (office room, library, laboratory, bedroom, lunch room, lounge, garage) in order to explore the effect of empty space, two human movement (walking towards each other and random movement) on Wi-Fi link throughput originated from file sharing between two wireless laptops concluding that the average throughput performance is $2.3\pm 0.3$ Mbps. Authors also reported that the noticeable throughput deviation ($\sim 0.3$ Mbps) is seen when two human movement is introduced rather than the empty space. But the throughput deviation due to different activities performed by two humans (walking towards each other vs random movement) is $\pm 0.07$ Mbps on average which is insignificant. 

This suggests that Wi-Fi packets don’t arrive after constant time interval, hence there exists variational packet-arrival-time-difference caused by multipath propagation effects. This also reminds that, detection of 13 two-human-interactions from Wi-Fi signal using AI is a very much arduous task due to negligible throughput deviation for different activities performed by two humans.

\subsection{Received Signal Strength Indicator (RSSI)}
\label{subsec:rssi}
Transmitted radio EM wave encounters fading, path loss, destructive interference during multipath propagation hence loses signal power over distance or attains power in case of noise addition and constructive interference. Received Signal Strength Indicator (RSSI) is a relative measurement of the signal power received at each receive antenna and the range of the RSSI value along with calculation depends on NIC manufacturer since there is no fixed standard for RSSI calculation. The RSSI value is observed to be greatly affected in different environments in a field study \cite{RSSI} on the characteristics of RSSI, especially with the presence of objects in signal path. Recently, RSSI has also been intended to be used for applications like human localization in indoor environment \cite{RSSIlocalization}. 

Therefore, RSSI values create unique variational pattern with the unique movement during different human-to-human interactions performed along signal LOS path that might be translated into meaningful classification using deep learning techniques.

\subsection{Automatic Gain Control (AGC)}
\label{subsec:agc}
Multipath propagation affects the transmitted signal amplitude (hence power) resulting to be either high, low, or just the exact at the receiving antenna. However, translatable-stable signal amplitude is an imperative for further processing of the signal using the receiver electronic circuits at which point the Automatic Gain Control (AGC) concept steps in. Delayed AGC is a feedback-controlled operational amplifier circuit including electronic switches (e.g. diode, relay etc.) aiming to dynamically adapt the output signal voltage to a set-point creating suitable DC voltage-bias using variable-gain-amplifier (VGA) and error-amplifier circuit-network so that the downstream circuit at the receiver chain gets a stable signal in the radio-frequency (RF) and intermediate-frequency (IF) stages irrespective of the varying received signal amplitude. Frequency bandwidth of the AGC circuit is defined by the operating bandwidth of the transmitter. \cite{AGC1, AGC2, AGC3}

AGC value (in dB) supplements the RSSI (in dB) deficit, that’s why kept as an important feature for the proposed deep-learning model.

\subsection{Channel State Information (CSI)}
\label{subsec:csi}
\begin{figure*}[ht]
\centering 
\resizebox{8cm}{!}{\includegraphics{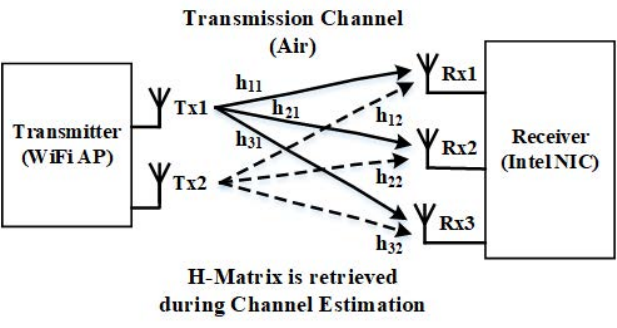}}
\caption{{Intuition of CSI array for single subcarrier}}
\label{fig: MIMO_CSI}
\end{figure*}

Transmitted analog EM wave reaches to the receiver passing through the overall spatial environment (e.g. free space) is referred to as transmission channel. Channel conditions and properties accumulate the overall combination of signal fading (reflection, absorption, interference, scattering, path loss, packet delay etc.) due to multipath propagation and unique human-to-human interactions along LOS path. Channel State Information (CSI) (Figure-\ref{fig: MIMO_CSI}) describes how the transmitted EM beam of each Wi-Fi packet passes through the channel encountering those effects and mimics the overall channel properties of a MIMO-OFDM system in the form of a (NTx, NRx, NSc)-sized array (H matrix) formed of complex elements describing magnitude and phase information of the channel impulse response which can be estimated both at the transmitter and receiver side using various channel estimation methods \cite{ChE1, ChE2}; here, NTx, NRx = Number of Transmit and Receive antennas and NSc = Number of Subcarriers used during OFDM modulation. 

This experiment used (2, 3, 30)-sized complex valued CSI array that was estimated at the receiver side using a publicly available CSI tool \cite{csi_tool}.

Each Wi-Fi packet is received within very small period of time after transmission, thus, quasi-static fading and a linear-time-invariant channel is assumed. If for a multipath channel, ${{\alpha }_{k}}$ be the channel attenuation over $k_{th}$ path, $d(\tau -{{\tau }_{k}})$ be the signal propagation delay along $k_{th}$ path at ${{\tau }_{k}}$ delay index of the channel impulse response, then each element $h_{T,R,S}$ of the H matrix for $T_{th}$ transmit antenna, $R_{th}$ receive antenna and $S_{th}$ subcarrier can be expressed as \cite{IEEE80211n, phdthesis, csi},

{\footnotesize \[\tag{8}{{h}_{T,R,S}}(\tau )=\sum\limits_{k=1}^{N}{{{\alpha }_{k}}{exp\left({-j\omega {{\tau }_{k}}}\right)}d(\tau -{{\tau }_{k}})\text{\;\;\; ; \; here,\;angular\;frequency,\;}\omega =2\pi {{f}_{subcarrier}}}\]}

\noindent CSI array for each Wi-Fi packet,  
\[\begin{array}{l}\tag{9}
H={{\left[ \begin{matrix}
   \left[ \begin{matrix}
   \left[ \begin{matrix}
   {{h}_{1,1,1}} & {{h}_{1,1,2}} & \ldots  & {{h}_{1,1,30}}  \\
\end{matrix} \right]  \\ \\
   \left[ \begin{matrix}
   {{h}_{1,2,1}} & {{h}_{1,2,2}} & \ldots  & {{h}_{1,2,30}}  \\
\end{matrix} \right]  \\ \\
   \left[ \begin{matrix}
   {{h}_{1,3,1}} & {{h}_{1,3,2}} & \ldots  & {{h}_{1,3,30}}  \\
\end{matrix} \right]  \\ \\
\end{matrix} \right]  \\ \\
   \left[ \begin{matrix}
   \left[ \begin{matrix}
   {{h}_{2,1,1}} & {{h}_{2,1,2}} & \ldots  & {{h}_{2,1,30}}  \\
\end{matrix} \right]  \\ \\
   \left[ \begin{matrix}
   {{h}_{2,2,1}} & {{h}_{2,2,2}} & \ldots  & {{h}_{2,2,30}}  \\
\end{matrix} \right]  \\ \\
   \left[ \begin{matrix}
   {{h}_{2,3,1}} & {{h}_{2,3,2}} & \ldots  & {{h}_{2,3,30}}  \\
\end{matrix} \right]  \\
\end{matrix} \right]  \\
\end{matrix} \right]}_{\left( NTx,\;NRx,\;NSc \right)}}
\end{array}\]

\vspace{5mm}
\noindent Received sequence for each Wi-Fi packet at the receiving antennas,
{\footnotesize \[\begin{array}{l}\tag{10}
 {{\left[ Y \right]}_{\left( NRx,\;NSc \right)}}=\sum\limits_{sum\_along\_Tx}{\left( {{\left[ H \right]}_{\left( NTx,\;NRx,\;NSc \right)}}*{{\left[ X \right]}_{\left( NTx,\;NSc \right)}} \right)}+{{\left[ AWGN \right]}_{\left( NRx,\;NSc \right)}} \\ \\
 
 \;\;={{\left[ \begin{matrix}
   \left[ \begin{matrix}
   {{h}_{1,1,1}}{{x}_{1,1}}+{{h}_{2,1,1}}{{x}_{2,1}} & {{h}_{1,1,2}}{{x}_{1,2}}+{{h}_{2,1,2}}{{x}_{2,2}} & \ldots  & {{h}_{1,1,30}}{{x}_{1,30}}+{{h}_{2,1,30}}{{x}_{2,30}}  \\
\end{matrix} \right]  \\ \\
   \left[ \begin{matrix}
   {{h}_{1,2,1}}{{x}_{1,1}}+{{h}_{2,2,1}}{{x}_{2,1}} & {{h}_{1,2,2}}{{x}_{1,2}}+{{h}_{2,2,2}}{{x}_{2,2}} & \ldots  & {{h}_{1,2,30}}{{x}_{1,30}}+{{h}_{2,2,30}}{{x}_{2,30}}  \\
\end{matrix} \right]  \\ \\
   \left[ \begin{matrix}
   {{h}_{1,3,1}}{{x}_{1,1}}+{{h}_{2,3,1}}{{x}_{2,1}} & {{h}_{1,3,2}}{{x}_{1,2}}+{{h}_{2,3,2}}{{x}_{2,2}} & \ldots  & {{h}_{1,3,30}}  \\
\end{matrix}{{x}_{1,30}}+{{h}_{2,3,30}}{{x}_{2,30}} \right]  \\
\end{matrix} \right]}_{\left( NRx,\;NSc \right)}} \\ \\

\;\;\;\;\;\;\;\;\;\;\;\;\;\;\;\;\;\;\;\;\;\;\;\;\;\;\;\;\;\;\;\;\;\;\;\;\;\;\;\;\;\;\;\;\;\;\;\;\;\;\;\;\;\;\;\;\;\;\;\;\;\;\;\;\;\;\;\;\;\;\;\;\;\;\;\;\;\;\;\;\;\;\;\;\;\;\;\;\;\;\;\;\;\;\;\;\;\;\;\;\;\;\;\;\;\;\;\;\; + {{\left[ AWGN \right]}_{\left( NRx,\;NSc \right)}}  
\end{array}\]}

In an indoor environment having static surrounding objects at fixed position but two humans performing different mutual interactions along LOS path creates idiosyncratic channel variation over the total time period that appear to be a classifiable pattern to the proposed deep learning model.

\section{Methodology and Framework}
\label{sec:methodology}

The human-to-human mutual interaction recognition from Wi-Fi data is a time-series classification problem and for that, this study proposes an auspicious Attention Bidirectional Gated Recurrent Neural Network Unit (Attention-BiGRU) model with executable Graphical User Interface (GUI) software implementation. The proposed solution utilizes WiFi signal noise, packet arrival time difference, RSSI value of three receive antennas, AGC and CSI array data features for thirteen distinct human-to-human mutual interaction classification performed in an indoor environment within the line-of-sight Wi-Fi coverage area of the transmitted signal. The cognitive ability of the developed GUI software solution possesses significant potential for indoor activity monitoring, surveillance system, smart health monitoring systems and independent assisted living because the software can alongside plot and save the time-series data classification results with a satisfactory accuracy having minimal interaction-transition region classification errors. The overall project structure is illustrated as block diagram in the Figure-\ref{fig: BlockDiagram}.

\subsection{Data Pre-processing}
\label{subsec:dat_preproc}

\begin{figure*}[htp]
\centering 
\resizebox{\textwidth}{!}{\includegraphics{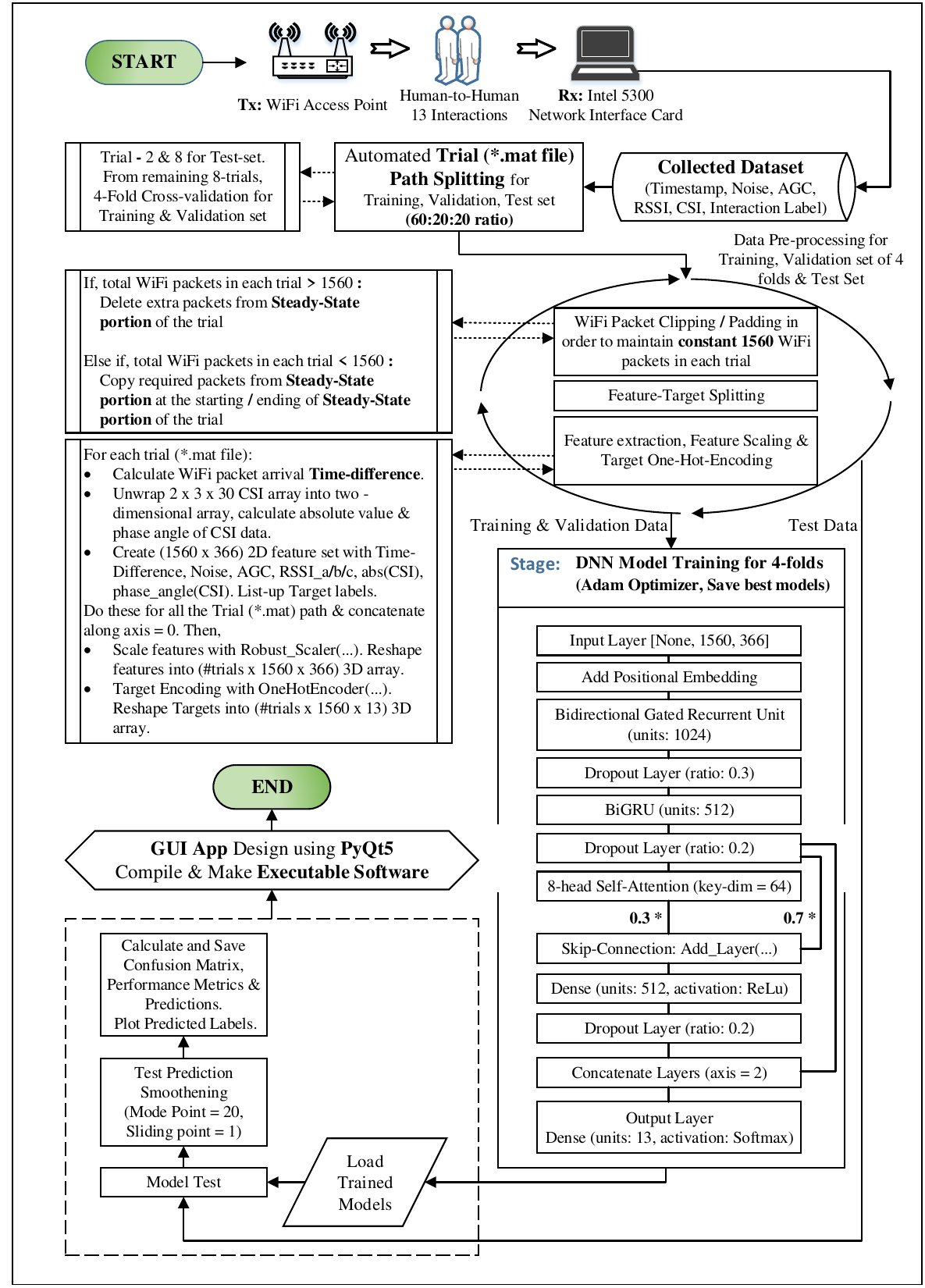}} 
\caption{{Overall Project Structure}}
\label{fig: BlockDiagram}
\end{figure*}

During data pre-processing, data were split into $60:20:20$ ratio for training-validation-test set. The algorithm at first detects the number of available Wi-Fi packets of a given sequence. Unequal length of data sequences are made equal to 1560 Wi-Fi packets by either clipping or padding of required samples because the deep learning model requires a fixed number of input data size. Then feature extraction is performed according to Figure-\ref{fig: BlockDiagram}. Packet Arrival Time Difference feature is calculated from the Time-Stamp column of the original dataset. CSI array of each Wi-Fi packet of the original dataset is of $2\times3\times30$ size which is unwrapped into a 2D array, then magnitude and phase angle is calculated. Other feature values (Time Difference, Noise, AGC, RSSI of three receive antennas) are concatenated alongside the CSI array magnitude and phase. This is done for all 1560 Wi-Fi packets of each *.mat file which produces $1560\times366$ 2D array in *.csv format for each *.mat file.   

The array is then undergone through Robust Scaler for standardization. Deep learning models exhibit dire performance on snaggy dataset due to outliers present in the data-features or skewed data-distribution that demands for methodical data standardization during data pre-processing step. Robust scaler takes care of the outliers present in individual feature vector by removing median, scales the values within specific range and centers-up the vector to form Gaussian distribution maneuvering $25^{th}$ to $75^{th}$ interquartile range \cite{sklearn}. Target values are encoded with One Hot Encoder for making suitable for further processing.

\subsection{Attention-BiGRU Model Design, Training and Test}
\label{subsec:model_train_test}

With the pre-processed data, model training was performed using the proposed Attention-BiGRU deep learning architecture. The proposed deep-learning architecture is a functional-type architecture comprised of an input layer, a positional embedding adder layer, two BiGRU layers having 1024 and 512 units respectively, one self-attention layer, one skip connection adder layer, two dense layers having 512 and 13 units, three dropout layers having 0.3 and 0.2 dropout ratios in order  to stay clear of overfitting issue, and one concatenation layer. Followed by the BiGRU operations, the under-processing array is fed to a Self-Attention layer having 8-heads with 64 key-dimension size. $30\%$ weighted self-attention values are added to the $70\%$ weighted previous-to-attention-values in the skip connection layer. This array is then passed through a dense layer and then the array is concatenated with the previous-to-attention layer values. Finally, this array is passed through the output dense layer having Softmax activation function that produces the time-series classification result. Whether GRU should be used as the Recurrent Neural Network, whether GRU layers should be bi-directional or uni-directional, how many BiGRU and dense layers need to be used, how many units of those layers need to be used, how many (multi-)heads and key-dimension of the self-attention layer should be used, how much weight should be given to the input threads of skip-connection layer, the arrangement and accurate placement of all the layers one after another are determined through trial and error process. The best combination is chosen for the finally proposed deep-learning model which had given the best possible outcome.

It is stated in Dataset Description (Section-\ref{sec:data_desc}) that trial files for different interaction classes have time-length inequalities. Primarily, they were made of equal length by either clipping or padding required WiFi packet data from/to the steady-state portion during data pre-processing step. But, each trial file consists of two interactions (steady-state and another out of remaining 12 mutual-interactions) merged together. Steady-state interaction was performed either at the beginning or at the last portion of each trial. The 13 mutual human-to-human interactions performed by various subject pairs had different height, weight and might have slightly different body-posture for a specific interaction which contributed micro-variation in CSI array for that interaction. Again, clipping/padding steady-state portion solved the overall trial file length inequality problem but still the time-length of steady-state interaction remains unequal. All of these challenges are mitigated by using Positional Embedding, Self-Attention, Skip Connection, and Custom Test Prediction Smoother layers. BiGRU and Dense layers mainly took care of the task of time-series classification. 

\vspace{8pt}
\noindent Basic layers used in the proposed deep learning Attention-BiGRU model are described as follows.

\subsubsection{Positional Embedding}
\label{subsec:pos_embd}

To deal with the interaction-transition region and satisfactory persistent labeling of the timeseries samples, it is essential to attain some essence of positional knowledge \cite{posembd} by the trained model for learning body-posture sequence-order since the dataset comprises interactions of variable data acquisition time-length and each interaction might have small chunk of body-posture correlating another interaction (e.g. `Punching/Pointing with left/right hand’ has similar state until just before punched/pointed-with-finger). Positional embedding is frequently being used in transformer-based models in natural language processing. However, such embedding layer also shows promising aspects in case of Attention-RNN models latterly in trend, as well, explored through this experiment.

\subsubsection{Bidirectional Gated Recurrent Unit (BiGRU)}
\label{subsec:bigru}
Analysis of time-bounded 1560 Wi-Fi packets per trial falls under sequence learning task and Gated Recurrent Neural Network Unit (GRU) \cite{gru} is scrutinized to be performing best in this experiment compared to Long-Short-Term-Memory (LSTM) which is another early variant of Recurrent Neural Network. 

Being neoteric, GRU offers some computational advantages over Vanilla-RNN and LSTM. While learning long term dependencies, GRU resolved the issue of vanishing and exploding gradient problem faced by Vanilla-RNN \cite{grufin}. On top of that, GRU incorporates fewer computational-states with less gates that provide faster computation with reduced complexity compared to LSTM. Each GRU unit intakes particular input along with previous hidden state corresponding to time-sample $t$, finds out whether the input is a potential hidden-state to be stored by using hyperbolic tangent and sigmoid activation function in candidate hidden state and reset gate thus serving as short-term-memory, and finally produces the current hidden state value (also final output value at the end) through reckoning of short-term-memory information from candidate state with previous hidden state value by using another sigmoid activation function in update gate thus serving as long-term-memory. On the whole, each GRU unit has two gates which maintain only the hidden state. The hidden-state at time $t$ of a fully-gated GRU unit can be expressed as,
\noindent {\footnotesize
\[\begin{array}{l}\tag{11} H\left( t \right)=H\left( t-1 \right)\odot {{\left[ 1+{exp\left(\alpha\right)} \right]}^{-1}}+\left[ 2{{\left[ 1+{exp\left(-2\beta\right)} \right]}^{-1}}-1 \right]\odot{{\left[ 1+{exp\left(-\alpha\right)} \right]}^{-1}}; \\ \\
 \text{here,\;\;}  \alpha = {W_{1}^{u}x\left( t \right)+W_{2}^{u}H\left( t-1 \right)+{{B}^{u}}} \\ 
 \text{\;\;\;\;\;\;\;\;\;} \beta = {W_{1}^{h}x\left( t \right)+W_{2}^{h}\left[ H\left( t-1 \right)\odot {{\left[ 1+{exp\left(-\gamma \right)} \right]}^{-1}} \right]+{{B}^{h}}}\\
 \text{\;\;\;\;\;\;\;\;\;} \gamma = {W_{1}^{r}x\left( t \right)+W_{2}^{r}H\left( t-1 \right)+{{B}^{r}}}\\ \\
 \text{\;\;\;\;\;\;\;\; x = input, t = time, u = update gate, r = reset gate, h = candidate hidden state, } \\ 
 \text{\;\;\;\;\;\;\;\; H = hidden state value, } 
 \text{W}_{n}^{g}\text{ = }{{\text{n}}_{th}}\text{ weight matrix of gate/state g, }\\
\text{\;\;\;\;\;\;\;\;\;}{{\text{B}}^{g}}\text{ = bias of gate/state g, }\odot \text{ = Hadamard Product}
\end{array}\]}

Bidirectional \cite{bidirectional} operation of such sequence-processing unit accumulates both the preceding hidden-state information during forward propagation and succeeding hidden-state information during backward propagation so as to produce single output for the corresponding single input, and pledges more accurate output than unidirectional layer for the dataset of this experiment since information from both past and future is preserved. Bidirectional outputs for default backward layer were being concatenated in this experiment, however, output for other available merge-modes and custom backward layer as well are yet to be explored.

\subsubsection{Multi-Head Self Attention}
\label{subsec:attention}
Self attention \cite{attention} is a scoring mechanism for each sample of an input sequence with respect to that particular sequence based on a context in which input samples interrelate with each other using Query, Key, Value matrices with a view to finding out valuable portions and give attention score to each sample accordingly. Multi-Head (say, number of heads = N) Self Attention gives the opportunity to score based on N representation contexts or subspaces using N-sets of Query, Key, Value matrices and final scores from each head are concatenated together. If X be the input array to an attention layer and W be the weight matrices, then Multi-Head Self Attention Score can be expressed as,
{\footnotesize \[\begin{array}{l}\tag{12}\text{Score}=\left[ {{W}_{Final}} \right]\times \underset{\text{for Head = 1\;to\; }\!\!\#\!\!\text{ Heads}}{\mathop{\text{Concatenate}}}\,\left( \frac{{exp\left(\alpha\right)}}{\sum\limits_{\text{array element}=1}^{N}{{exp\left(\alpha\right)}}}\times \left[ \left[ X \right]\times \left[ W_{Value}^{Head} \right] \right] \right) \text{\; ;} \\ \\
\text{here, \;} \alpha = {\frac{\left[ \left[ X \right]\times \left[ W_{Query}^{Head} \right] \right]\times {{\left[ \left[ X \right]\times \left[ W_{Key}^{Head} \right] \right]}^{Transpose}}}{\sqrt{\text{Key-array dimension}}}}
\end{array}\]}

\subsubsection{Skip Connection}
\label{subsec:skip_connection}
In the proposed model, 30\% of each array element of the multi-head self-attention score is added to the 70\% of that particular corresponding element of the input array to attention layer for which the score is calculated. In other words, attention score array is given 30\% weight, input array to attention layer is given 70\% weight, and then added together. 

\subsubsection{Dense Layer}
\label{subsec:dense}
A dense layer consists of a number of nodes, also known as neurons, wherein each node takes input from all or most (controlled by dropout ratio) of the nodes of the preceding neural network layer for calculation of probabilities using specified activation function.

\subsubsection{Dropout Layer}
\label{subsec:dropout}
To control the data overfitting, dropout layer randomly disconnects certain nodes of a neural network layer by setting input values to 0 during each step of the model training procedure. Percentage of dropped out node is controlled by the user defined parameter dropout-ratio. Thus, the weights are trained to be immune to overfitting. However, this layer only works during model training but stays dormant during model test-phase and in application.

\subsubsection{Adaptive Moment Estimation (Adam) Optimizer}
\label{subsec:Adam} 
Upon establishing the fundamental architecture of a deep learning model, the subsequent step involves optimizing the model. This entails defining a loss function and implementing optimization algorithms to adjust the model's activation function weights and hyperparameters to minimize the loss function. The iterative optimization process persists until the loss function converges to the global minimum, yielding a highly accurate and precise model. The optimizer algorithm employed dictates the learning time required for the iterations and determines the optimal weights and parameters. A recent study by Hassan et al. \cite{adambest} has found that the Adaptive Moment Estimation (Adam) Optimizer outperforms several other built-in optimization algorithms of the Python Keras library for Deep Neural Networks. Consequently, the authors of this article have incorporated the Adam optimizer as the Mini-Batch Gradient Descent Optimization algorithm for the experiments of mutual human-to-human interaction recognition in this study.

According to Hassan et al. \cite{adambest}, Kingma et al. \cite{adamoriginal}, and Ajagekar \cite{adameasy}, the Adam optimizer is a modified and combined version of two Stochastic Gradient Descent methods: Momentum and Root Mean Square Propagation. Adam calculates the momentum for each parameter, which involves using estimations of the first and second moments of the gradient to adapt the learning rate for each weight in the neural network. Subsequently, Adam optimizer updates the weights through successive iterations based on the algorithm described in the articles by Kingma et al. \cite{adamoriginal} and Ajagekar \cite{adameasy}.

\subsubsection{Custom Test Prediction Smoother}
\label{subsec:smoother}
Few very-small chunks of oscillatory predictions were being observed in the test prediction that was changing within nano-seconds especially during interaction transition and other parts. But human-to-human interaction cannot change abruptly within nanoseconds. A custom smoother was designed using 1 moving point and 20 mode point in order to rectify the problem. If the mode of the classifications of 20 sample-points preceding and 20 sample-points succeeding to a single sample-point is similar, then the classification of that single sample-point is set equal to the mode classification; if not, then left as it is.

\vspace{10pt}
\noindent The overall training is performed through 4-Fold cross-validation and thus four trained model weights have been generated and saved. During Test-phase, each of the test *.mat file is classified using all those four trained model weights separately and the mode of the classification of each Wi-Fi packet (i.e. time-stamp) is considered as the classified interaction. Then again, the classified array is rectified using a custom Test Prediction Smoothening process having 20 mode point and 1 sliding point. This classified array is considered as the final time-series classification of 1560 Wi-Fi packets of each data *.mat file. Thus, the highly accurate and minimal interaction-transition region error (Figure-\ref{fig: GUI}) is ensured by the proposed solution. The model was optimized using Adam optimizer, 300 epochs with early stopping criterion and mini-batch size 12. Reliable learning rate was adapted using ReduceLROnPlateau function. 

\vspace{8pt}
\noindent It is to be mentioned that, the performance metrics (accuracy, precision, recall, loss) reported for training are obtained from model training and test-loss during model test using Keras evaluation metrics but the performance metrics (accuracy, precision, recall, F1-score) reported for test are obtained after test prediction smoothening and then using weighted average Scikit-learn metrics.

\subsection{Executable Graphical User Interface Software Design}
\label{subsec:gui_app}

To make the proposed solution user-friendly, an executable (not standalone) graphical user interface (GUI) software has been developed using PyQt5 \cite{pyqt5} GUI app builder Python module so that the user can provide particular input file to the software and within a while can get the classified results directly plotted on-screen in a time-series fashion, the classification results are being saved to the computer simultaneously for later use. With this executable software, a user can classify either a particular trial file or an entire folder containing multiple trial files that will be fetched by QFileDialog box and the selected directory-path will be displayed on a text-label. The application is designed such that it will search for how many trained models (*.h5 file) are present in a folder named `cloverleaf’, use all of them and calculate the mode during classification of the trial file. This gives full control over the user to download and use as many trained models as desired (higher, the better). The software can automatically detect and activate the GPU (if present) of the user device for faster calculation, however, it will calculate at a slower speed if GPU is missing in the device. The procedural progress will simultaneously be displayed on a progress bar and on an inactive push-button label. Following completion, the classified trials will be saved as a *.csv file in a datetime-bearing folder in the working directory, the path of which will be displayed on a text label. Finally, the classified trials will be plotted on QtWebEngine using the Plotly Offline python module and the plots can be switched back and forth by rotating a knob on the right-screen. A user notification will also appear at the bottom of the screen with instructions on graphics control. The program is made capable to scan the data file for the true target labels, plot them alongside the classified labels, and omit true label plots if they are not found.

\vspace{0.6cm}
\noindent Model training and test-evaluation were performed using Google Colaboratory free plan (used scipy v1.7.2 and TensorFlow framework) but GUI software was designed and evaluated using personal laptop having following configurations: $8^{th}$ Generation Intel\textsuperscript{\textregistered} Core\textsuperscript{TM} i5-8250U processor, 8GB DDR4 RAM, 4GB DDR3 NVIDIA GeForce 940MX CUDA-enabled graphics card, Python v3.9.3 with PyCharm Community Edition v2021.3 IDE running on Windows 10 operating system.

\section{Result and Discussion}
\label{sec:result}

The contribution to this human-to-human mutual interaction recognition study is structured within three experiments. 

\vspace{2mm}

\begin{itemize}[noitemsep, topsep=0pt, leftmargin=*]
    \item[i.] The first experiment is focused on determining whether and to what extent the Wi-Fi signal along with channel parameters used in analysis are capable of mutual interaction recognition via the proposed deep-learning model. Experiment-1 involves training and model evaluation using the first pair of subjects (S1-S47), i.e. single folder containing 120 trials.
    
    \item[ii.] Micro-variation is introduced to the feature pattern describing similar interaction in case of multiple subject-pairs because of their diversified height, weight, gesture-style and clothing. Such physiological varieties are the source of slightly different pattern of body-reflected signal for the similar interaction performed by different subject-pairs that results in the aforementioned micro-variation of the feature pattern. The second experiment explores whether the proposed model can withstand these micro-variational pattern changes and whether the model gives satisfactory performance on test-data involving multiple subject-pairs if the model is properly trained with data from those specific subject-pairs. Experiment-2 is conducted on the first-ten pair of subjects (S1-S47 to S16-S41), i.e. 10 folders containing 1200 trials.
    
    \item[iii.] To explore the pertinent obstructions and the usage feasibility of the trained models obtained from experiment-2 in case of an unknown situation, a cross-test is performed in experiment-3 on the second-ten pair of subjects (S18-S57 to S34-S30).
\end{itemize} 

\vspace{2mm}

\begin{figure*}[htbp]
\centering 

\resizebox{9.4cm}{!}{\includegraphics{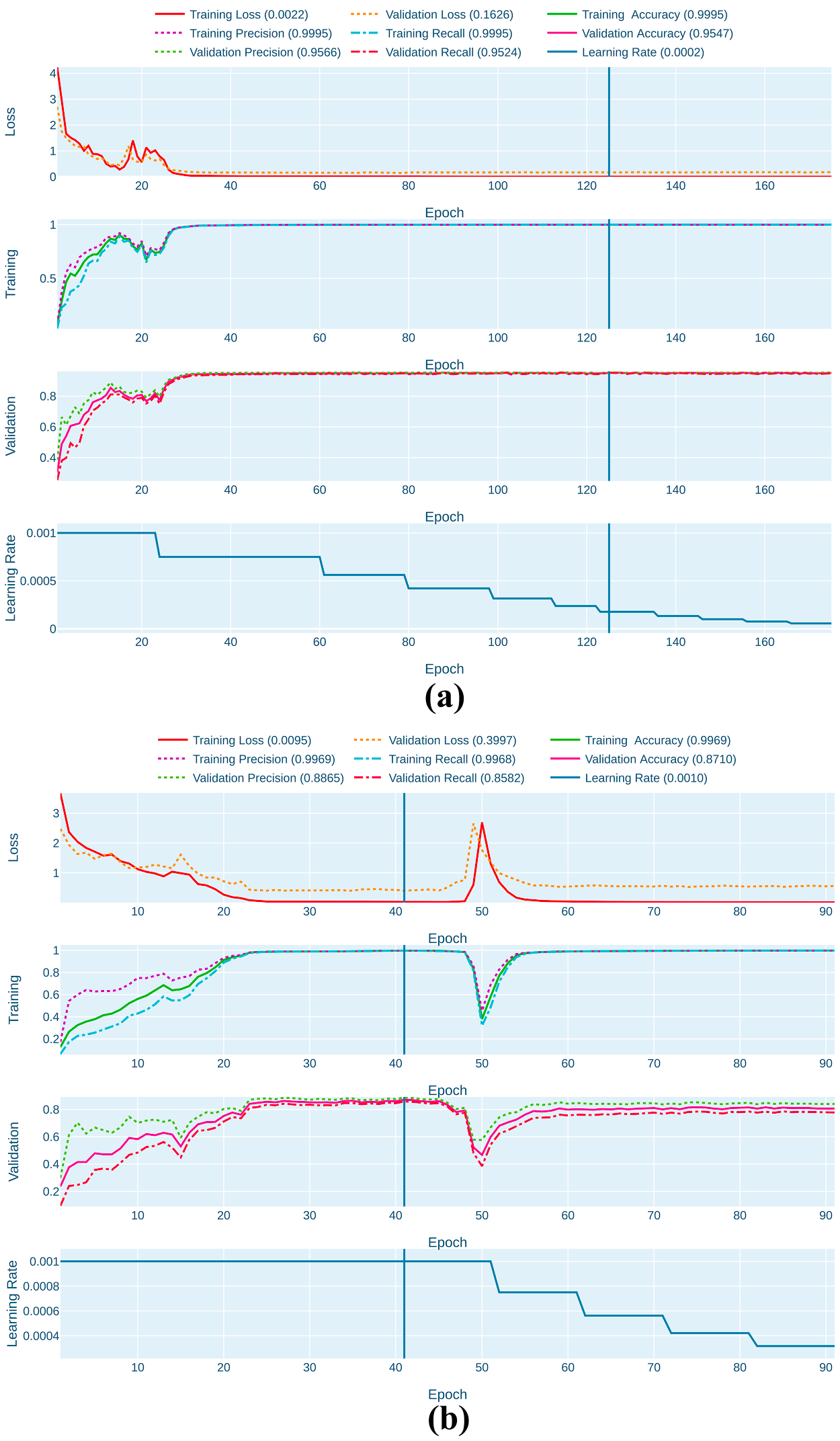}}

\caption{{Training and cross-validation performance: (a) Fold-3 of Experiment-1, (b) Fold-2 of Experiment-2. Vertical line and legend values correspond to the best epoch based on the best validation accuracy. Learning performance of the remaining folds can be viewed from \textbf{Supplementary Materials}}}
\label{fig: learningperformance}
\end{figure*}

\begin{figure*}[htbp]
\centering 

\resizebox{8.7cm}{!}{\includegraphics{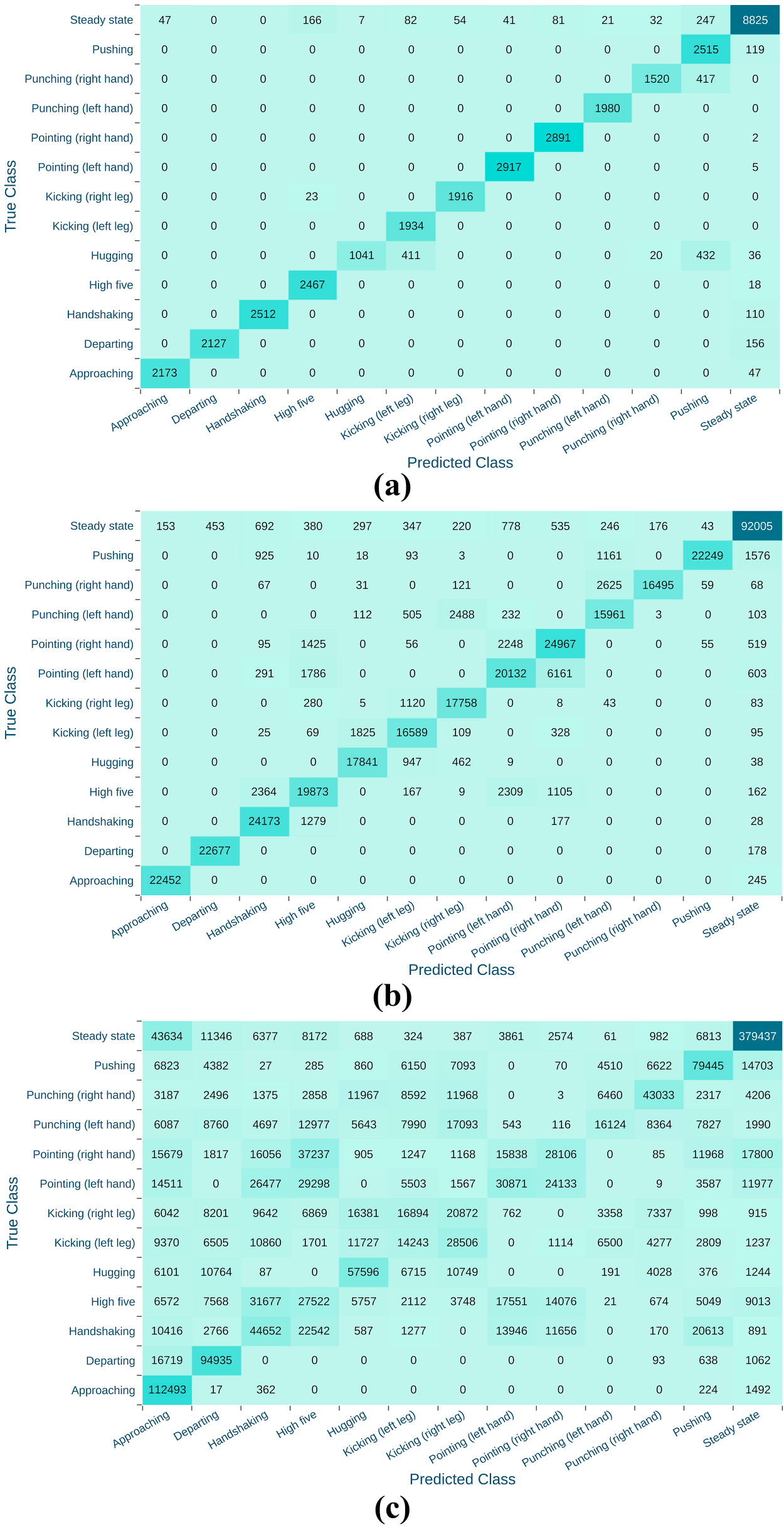}}

\caption{{Confusion matrix after test evaluation (a) Experiment-1: Mode Prediction of 4 Folds on 24 trial test-files, (b) Experiment-2: Mode Prediction of 4 Folds on 240 trial test-files, (c) Experiment-3 on new 1200 trial test-files. Remaining confusion matrices can be viewed from \textbf{Supplementary Materials}}}
\label{fig: confusionmatrix}
\end{figure*}

\noindent All the three experiments are set to be involved in classifying thirteen distinct mutual interaction target-classes. Learning curves obtained during the proposed deep-learning model training-phase are depicted in Figure-\ref{fig: learningperformance}, the confusion matrices are portrayed in Figure-\ref{fig: confusionmatrix}, and the performance metrics of the three experiments are tabulated in Table-\ref{tab: PerformanceMatrices}.

\vspace{4pt}
\noindent \textbf{Experiment 1:} It is observed that, whilst in training with $60:20$ $training:validation$ data for four folds, the model performed best on the third fold of cross-validation data showcasing more than 95\% performance metrics in case of single human-pair interaction, around 90\% performance metrics were being observed for another two folds, and the minimal performance metrics were found to be around 80\%. Trained models of all the remaining folds exhibited more than 90\% test benchmark on the remaining 20\% data except for the poor-performing fold which showed 88\% metric-value in the test-phase of experiment-1. Generalized classification from the mode of the predicted targets in the test-phase using the four trained models achieved around 94\% test benchmark which is the best in experiment-1 and also identical to its second fold. Confusion matrix Figure-\ref{fig: confusionmatrix}(a) also validates the underlying reason behind such high performance. Misclassifications observed in case of `steady-state’ are especially around the transition points from steady-state to another state or vice versa in a particular trial since Wi-Fi packets arrive within nanoseconds and thus the model gets blindfolded around the transition-points that produces swift dangling classification for few vicinal Wi-Fi packets. Apart from that there is almost no misclassification as can be verified from the confusion matrix of experiment-1. These findings prompted the study to continue investigation for the higher number of subject-pairs that the available RAM could accommodate.

\vspace{4pt}
\noindent \textbf{Experiment 2:} On the next step for the experiment-2, ten subject-pairs had been considered at once and further processing was carried out in the same way as experiment-1. Cross-validation performance in this case was noticed to be ranging between 80\% to 88\%, whereas the second fold of data performed the best with 87.1\% accuracy, 88.65\% precision, 85.82\% recall, and sparse categorical cross-entropy loss was minimized to 0.3997. Even though the trained model from third-fold data achieved the best individual test benchmark (86.5\%), generalized classification via mode-calculation of the predicted targets from four trained models outperformed each of the individual test performance metrics manifested with 88.3\% accuracy, 88.7\% precision, 88.3\% recall and 88.4\% F1-score. It is quite remarkable that inclusion of human-to-human interaction data performed by ten subject-pairs reached 88.5\% test benchmark which is relatively near to that for single human-pair. The test classification plot reveals that after training with huge amount of data from ten subject-pairs, the rapid oscillatory classification around transition points reduced noticeably to a trifling amount compared to experiment-1. A small number of misclassifications on few particular classes except steady state as seen from confusion matrix Figure-\ref{fig: confusionmatrix}(b) are due to either partial or infrequent complete misclassification of a particular trial. It is worth noting that the misclassified interactions have certain portions that are quite symmetrical to one another. For example, pointing or punching with right/left hand and high-five produces similar type of signal reflections from human body starting at the beginning of the interaction until just before pointed or punched either with right or left hand or given high-five. 


\begin{table*}[t]
\caption{Performance metrics for three experiments conducted on the whole}
\centering
\begin{adjustbox}{width=\textwidth}

{\begin{tabular}{cccccccccccc} \toprule
\multicolumn{12}{l}{\textbf{Experiment-1: Single Pair (S1-S47)}}  \\ \midrule
\multirow{2}{*}{Fold} & & \multicolumn{4}{c}{Cross-Validation Set} & &  \multicolumn{5}{c}{Test Set}   \\ \cmidrule{3-6} \cmidrule{8-12}
& & Accuracy & Precision & Recall & Loss & & Accuracy & Precision & Recall & F1-Score & Loss   \\ \midrule
1 && 0.8018 & 0.8143 & 0.7965 & 0.9837 && 0.883 & 0.905 & 0.883 & 0.880 & 0.716 \\
2 && 0.9365 & 0.9413 & 0.9335 & 0.2637 && \textbf{0.937} & 0.943 & \textbf{0.937} & \textbf{0.935} & \textbf{0.259} \\
3 && \textbf{0.9547} & \textbf{0.9566} & \textbf{0.9524} & \textbf{0.1626} && 0.913 & 0.920 & 0.913 & 0.912 & 0.412 \\
4 && 0.8984 & 0.9079 & 0.8939 & 0.3782 && 0.911 & 0.926 & 0.911 & 0.909 & 0.504 \\
Mode$^*$ && -- & -- & -- & -- && 0.936 & \textbf{0.944} & 0.936 & \textbf{0.935} & -- \\ \midrule \\ \midrule

\multicolumn{12}{l}{\textbf{Experiment-2: First Ten Pairs (S1-S47 to S16-S41)}}  \\ \midrule
\multirow{2}{*}{Fold} & & \multicolumn{4}{c}{Cross-Validation Set} & &  \multicolumn{5}{c}{Test Set}   \\ \cmidrule{3-6} \cmidrule{8-12}
& & Accuracy & Precision & Recall & Loss & & Accuracy & Precision & Recall & F1-Score & Loss   \\ \midrule
1 && 0.8030 & 0.8179 & 0.7927 & 0.7288 && 0.830 & 0.835 & 0.830 & 0.830 & 0.635 \\
2 && \textbf{0.8710} & \textbf{0.8865} & \textbf{0.8582} & \textbf{0.3997} && 0.858 & 0.862 & 0.858 & 0.859 & 0.452 \\
3 && 0.8601 & 0.8793 & 0.8393 & 0.4090 && 0.865 & 0.872 & 0.865 & 0.866 & \textbf{0.422} \\
4 && 0.8258 & 0.8407 & 0.8146 & 0.6439 && 0.829 & 0.832 & 0.829 & 0.828 & 0.607 \\
Mode$^*$ && -- & -- & -- & -- && \textbf{0.883} & \textbf{0.887} & \textbf{0.883} & \textbf{0.884} & -- \\ \midrule \\ \midrule

\multicolumn{12}{l}{\textbf{Experiment-3: Cross Test. Saved Models of four folds from Experiment-2 used to Test Whole Second}} \\
&& \multicolumn{10}{l}{\textbf{Ten Pairs (S18-S57 to S34-S30).}}  \\ \midrule
\multirow{2}{*}{Fold} & & \multicolumn{4}{c}{Cross-Validation Set} & &  \multicolumn{5}{c}{Test Set}   \\ \cmidrule{3-6} \cmidrule{8-12}
& & Accuracy & Precision & Recall & Loss & & Accuracy & Precision & Recall & F1-Score & Loss   \\ \midrule
Mode$^*$ && -- & -- & -- & -- && 0.506 & 0.501 & 0.506 & 0.489 & -- \\

\bottomrule
\multicolumn{12}{r}{\small *Classified using the four trained models and used their `Mode' as final classified label}
\end{tabular}}

\end{adjustbox}
\label{tab: PerformanceMatrices}
\end{table*}

\vspace{4pt}
\noindent \textbf{Experiment 3:} So far, everything has worked out well because we are able to classify time-sequence data and can even observe for how long the human-to-human interaction had been performed. Now comes the difficult part. Is the trained deep learning model capable of doing well on completely new subject-pairs for which it has never been trained? If not, what are the issues that the next research perspective should address? To reach these points, experiment-3 was carried out as a cross-test on the second-ten pair of subjects (S18-S57 to S34-S30). The four trained models from experiment 2 on 720 trials were used to test on 1200 subsequent trials, however the outcome was not satisfactory showing 50\% test benchmark. The confusion matrix of experiment-3 is given in Figure-\ref{fig: confusionmatrix}(c), and performance metrics of the three experiments are tabulated in Table-\ref{tab: PerformanceMatrices}. 

According to Ashleibta et al. \cite{CSI5G} and Damodaran et al. \cite{ubuntucsi}, IEEE 802.11n compatible Wi-Fi access points employ 51 subcarriers while modulation, however the Intel 5300 NIC only reports 30. Channel State Information data of remaining 21 subcarriers are lost and hence 46\% of the data were not involved in analysis. Again, the whole dataset contains 4800 trial files for 40 distinct subject-pairs, but because this study was conducted using the Google Colaboratory free plan, only 720 trials (60\% training data of 1200 in-total) from 10 subject-pairs could be used for the proposed deep-learning model training at once due to RAM limits. As a result, a bulk of micro-variation of CSI patterns from different subject-pairs (caused by different subject-specific style of gestures and variational signal fading due to unique lossy human body medium for different height and weight) were missed during the training phase. On a broad scale, these could be the possible reasons behind Domain-Shift Problem \cite{domain_shift} encountered in experiment-3. To reduce the difficulties seen in experiment 3, various Domain-Adaptation \cite{domain_adaptation1, domain_adaptation2} strategies along with sophisticated signal pre-processing techniques and inclusion of CSI data for all 51 subcarriers should be incorporated in future study.

\begin{figure*}[htb]
\centering 
\resizebox{\textwidth}{!}{\includegraphics{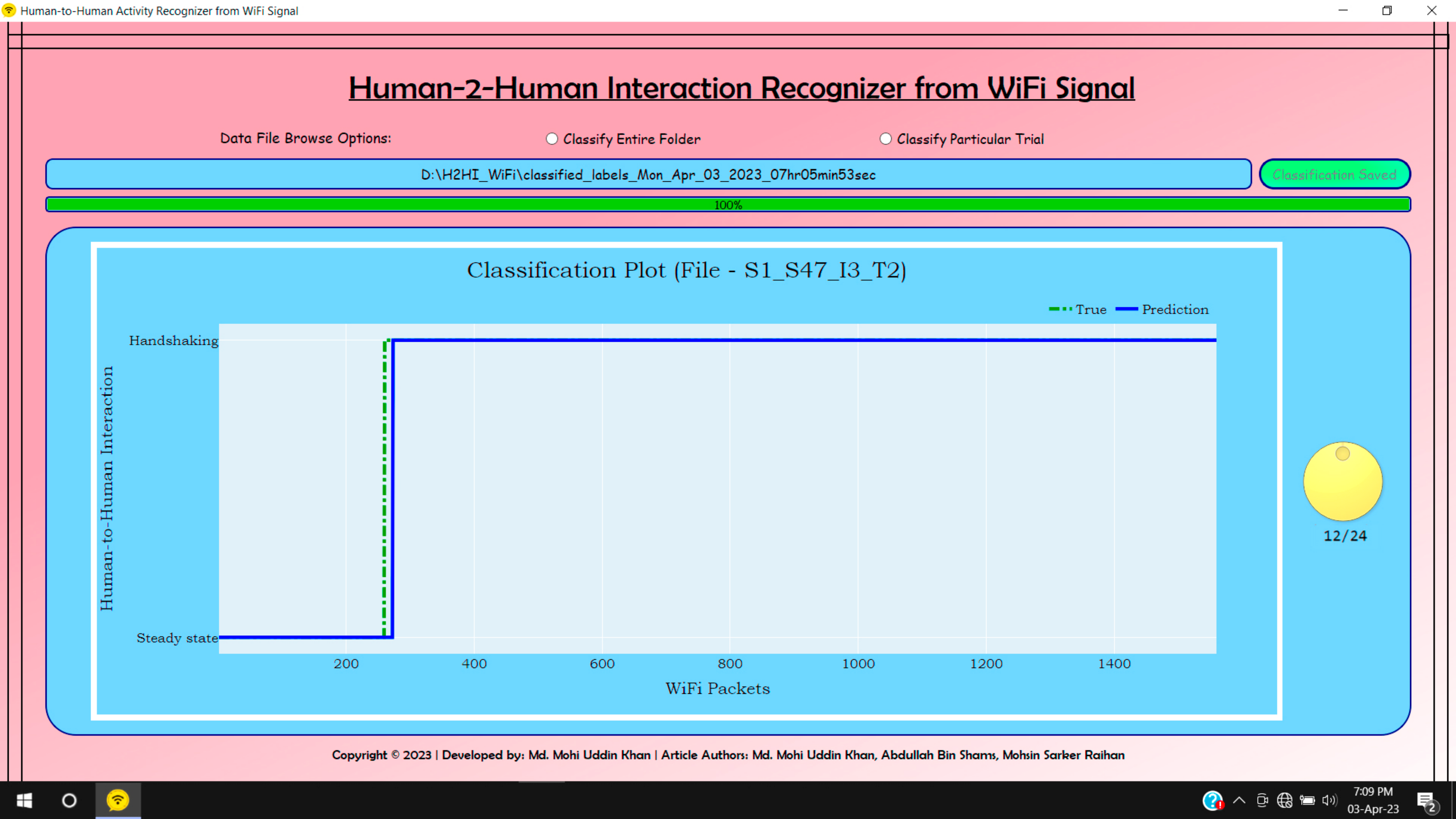}}
\caption{{Graphical User Interface (GUI) software designed using PyQt5. A video showcasing the software along with the software itself can be obtained from \textbf{Supplementary Materials}}}
\label{fig: GUI}
\end{figure*}

\vspace{4pt}
\noindent \textbf{Executable GUI Software Demonstration:} Experiment-2 demonstrates that the proposed classifier model can recognize human-to-human interactions with reasonable performance metrics having upto ten subject-pairs if the model is well-trained with the data regarding those specific subject-pairs. So in the end, a PyQt5 \cite{pyqt5} graphical user interface (GUI) software (Figure-\ref{fig: GUI}) was developed to conclude this study utilizing the trained models from experiment-2. A video showcasing the software demonstration along with the software itself can be obtained from \textbf{Supplementary Materials}.

\vspace{4pt}
\noindent \textbf{Comparative Analysis with State-of-the-Art Literature:} A comprehensive analysis and comparison of the proposed methodology with the state-of-the-art literature is presented in Tables \ref{tab: comptable1}-\ref{tab: comptable4}. The comparative analysis reveal that the majority of recent studies utilized the 5 GHz channel frequency and 20/40 MHz carrier bandwidth variant of the WiFi 802.11n communication system with Single Input Multiple Output (SIMO) radio links, while the alternative 2.4 GHz channel frequency and real-life scenario with Multiple Input Multiple Output (MIMO) radio links were left unexplored. Additionally, most studies employed two laptops or two WiFi routers as transmitter-receiver pairs, simulating radio communication, which is not practical for continuous everyday use. In contrast, this study utilized a MIMO radio link with a multi-antenna WiFi router as the transmitter and a multi-antenna Intel NIC mounted in a PC as the receiver, and explored the 2.4 GHz channel frequency having 20 MHz carrier bandwidth variant of the WiFi 802.11n communication system. However, the use of an Intel NIC at the receiver restricted the number of reported subcarriers to only 30, which could be increased by employing the 5G band (3.75 GHz channel frequency) WLAN communication using Universal Software Radio Peripheral (USRP) devices in MIMO mode.

\begin{sidewaystable}
\caption{A Comprehensive Survey and Comparative Analysis of Related Works to this Article}
\centering
\begin{adjustbox}{width=\smallestside-2in}
{\begin{tabular}{llllll}\toprule
\textbf{Article} &\textbf{WLAN type} &\textbf{Communication} &\textbf{Communication platform} &\textbf{Device configuration} \\
& &\textbf{Link} & & \\\midrule
Wang et al. \cite{ieeelstm1} &WiFi 802.11n &Single &2 Intel 5300 NIC mounted in 2 laptops used as 3 antenna Tx, 3 antenna Rx &Laptop: Dell Latitude D820 \\
&Channel frequency: 5 GHz &Input &No. of subcarriers transmitted: 30 &GPU: NVIDIA Geforce GTX 1060 \\
&Carrier bandwidth: 20 MHz &Multiple &No. of subcarriers received: 30 & \\
& &Output &CSI extractor: Linux 802.11n CSI Tool \cite{csi_tool} & \\
& & & & \\
Damodaran et al. \cite{ubuntucsi} &WiFi 802.11n &Single &2 Intel 5300 NIC mounted in 2 laptops used as 3 antenna Tx, 3 antenna Rx &Laptop: 2 Lenovo laptop \\
&Channel frequency: 5 GHz &Input &No. of subcarriers transmitted: 30 &64 Bit Ubuntu v14.04 LTS OS \\
&Carrier bandwidth: 20 MHz &Multiple &No. of subcarriers reported by Rx: 30 & \\
& &Output &CSI extractor: Linux 802.11n CSI Tool \cite{csi_tool} & \\
& & & & \\
Ashleibta et al. \cite{CSI5G} &WLAN (5G band) &Single &2 USRP-X300/310 devices (each having a single antenna) &Computer: Intel(R) Core (TM) i7-7700 3.60 GHz \\
&Channel frequency: 3.75 GHz &Input &connected to computers used as Tx and Rx &processors having a 16 GB RAM \\
& &Single &No. of subcarriers transmitted: 51 &Ubuntu 16.04 operating system. Gnu Radio \\
& &Output &No. of subcarriers reported by Rx: 51 &is used to communicate with USRPs. \\
& & &CSI extractor: Python coding & \\
& & & & \\
Shalaby et al. \cite{csiRev_cnnGRUattention} &WiFi 802.11n &Single &Tx: 3 antenna WiFi Router, Rx: Laptop with 3 antenna Intel 5300 NIC &Laptop with Ubuntu OS \\
& &Input &No. of subcarriers transmitted: $>30$ & \\
& &Multiple &No. of subcarriers reported by Rx: 30 & \\
& &Output &CSI extractor: Linux 802.11n CSI Tool \cite{csi_tool} & \\
& & & & \\
Yang et al. \cite{autofi} &WiFi 802.11n &Single &Tx (1 antenna used) - Rx (3 antenna used): two TPLink-N750 routers. &Computer with single NVIDIA RTX 2080Ti GPU \\
&Channel frequency: 5 GHz &Input &No. of subcarriers for each pair of antennas: 114 & \\
&Carrier bandwidth: 40 MHz &Multiple &CSI extractor: Atheros CSI Tool \cite{csi_tool_atheros} & \\
& &Output &[Note: Atheros tool can only report CSI amplitude.] & \\
& & & & \\
Bocus et al. \cite{slfspcsi} &WiFi 802.11n &Multiple &Tx-Rx: 3-antenna Intel 5300 NIC. &- \\
&Channel frequency: 5 GHz &Input &One-Tx and two-Rx at different positions to form multi-view CSI & \\
& &Multiple &No. of subcarriers: 30 & \\
& &Output & & \\
& & & & \\
Xu et al. \cite{contrst_csi} &WiFi 802.11n &Single I/P &Tx: 1 antenna used (total no. of antenna not reported), Rx: 3 antenna &Computer with Nvidia Quadro P6000 GPU \\
& &Multiple O/P &No. of subcarriers: 30 & \\
& & & & \\
Al-qaness \cite{tailorcsi} &WiFi 802.11n &Multiple &Tx: TP-LINK TL-WR842N WiFi router with 2 antenna &Lenovo laptop installed with Ubuntu 14.04 OS \\
& &Input &Rx: Laptop installed Intel 5300 NIC with 3 antenna & \\
& &Multiple &No. of subcarriers reported by Rx: 30 & \\
& &Output &CSI extractor: Linux 802.11n CSI Tool \cite{csi_tool} & \\
& & & & \\
This Article &WiFi 802.11n &Multiple &Tx: Sagemcom-2704 WiFi router with 2 antenna &Data Collection: \cite{dataset} \\
&Channel frequency: 2.4 GHz &Input &Rx: PC having Intel 5300 NIC with 3 antenna &HAR model development: Google \\
&Carrier bandwidth: 20 MHz &Multiple &No. of subcarriers reported by Rx: 30 &Colaboratory Free Plan with TPU \\
& &Output &CSI extractor: Linux 802.11n CSI Tool \cite{csi_tool} &Software deployment: HP laptop with- \\
& & & &8th Gen Intel Core i5-8250U processor \\
& & & &8GB DDR4 RAM \\
& & & &4GB DDR3 NVIDIA GeForce 940MX- \\
& & & &CUDA-enabled GPU \\
& & & &Windows 10 OS \\

\bottomrule

\end{tabular}}
\end{adjustbox}

\label{tab: comptable1}
\end{sidewaystable}

\begin{sidewaystable}
\caption{A Comprehensive Survey and Comparative Analysis of Related Works to this Article (contd.)}
\centering
\begin{adjustbox}{width=\smallestside-1in}
{\begin{tabular}{lllllll}\toprule
\textbf{Article} &\textbf{Target class} &\textbf{Mutuality} &\textbf{Concurrency} &\textbf{`Inter-activity} &\textbf{Activity} \\
\textbf{} &\textbf{} &\textbf{(Independent/} &\textbf{(Single/} &\textbf{variability'} &\textbf{data} \\
\textbf{} &\textbf{} &\textbf{Mutual} &\textbf{Concurrent} &\textbf{addressed?} &\textbf{time-length} \\
\textbf{} &\textbf{} &\textbf{activity)} &\textbf{activity)} &\textbf{} &\textbf{inequality} \\\midrule
Wang et al. \cite{ieeelstm1} &Target classes (8): walking, falling, running, &Independent &Single &No &Yes \\
&sitting, picking, pushing, waving, boxing & & & & \\
& & & & & \\
Damodaran et al. \cite{ubuntucsi} &Target classes (5): walking, sitting, standing, &Independent &Single &No &No \\
&running, empty & & & & \\
& & & & & \\
Ashleibta et al. \cite{CSI5G} &Target classes (16): Empty, 1/2/3/4 person &Independent &Single &Yes &No \\
&standing, 1/2/3/4 person sitting, 1 sit+1 stand, & & & & \\
&1 sit+2 stand, 2 sit+1 stand, 2 sit+2 stand, & & & & \\
&1 person walking, 1 walk+1 sit, 1 walk+2 sit & & & & \\
& & & & & \\
Shalaby et al. \cite{csiRev_cnnGRUattention} &Target classes (6): Lie down, Fall, Walk, Run, &Independent &Single &Yes &No \\
&Sit down, Stand up & & & & \\
& & & & & \\
Yang et al. \cite{autofi} &Target classes (8): up and down, left and &Independent &Single &Yes &No \\
&right, pull and push, clap, fist, circling, & & & & \\
&throw, zoom & & & & \\
& & & & & \\
Bocus et al. \cite{slfspcsi} &Target classes (6): lay down, sit, stand, &Independent &Single &Yes &No \\
&stand from the floor, walk, body rotate & & & & \\
& & & & & \\
Xu et al. \cite{contrst_csi} &Office Room Dataset (7 classes): lie down, fall, &Independent &Single &Yes &- \\
&pickup, run, sit down, stand up, and walk & & & & \\
&Falldefi Dataset (2 classes): Fall down, other & & & & \\
&Signfi Dataset: (276 classes) & & & & \\
& & & & & \\
Al-qaness \cite{tailorcsi} &Target classes (9): dodge, push, &Independent &Single &Yes &No \\
&strike, circle, punch, bowl, & & & & \\
&drag, pull, and kick & & & & \\
& & & & & \\
This Article &Target classes (13): &Mutual &Concurrent &Yes &Yes \\
&steady-state, approaching, departing, & & (Steady-state + & & \\
&handshaking, high-five, hugging, & & any other out  & & \\
&pushing, kicking (left-leg, right-leg), & &  of the remaining & & \\
&pointing (left-hand, right-hand), & & 12 activities) & & \\
&punching (left-hand, right-hand) & & & & \\

\bottomrule

\end{tabular}}
\end{adjustbox}

\label{tab: comptable2}
\end{sidewaystable}

\begin{sidewaystable}
\caption{A Comprehensive Survey and Comparative Analysis of Related Works to this Article (contd.)}
\centering
\begin{adjustbox}{width=\smallestside-3.5in}
{\begin{tabular}{llllll}\toprule
\textbf{Article} &\textbf{\# original} &\textbf{Original} &\textbf{\# training} &\textbf{Data-preprocessing} \\
\textbf{} &\textbf{dataset} &\textbf{dataset} &\textbf{and test} &\textbf{} \\
\textbf{} &\textbf{features} &\textbf{features} &\textbf{samples} &\textbf{} \\\midrule
Wang et al. \cite{ieeelstm1} &1 &CSI array shape per sample &Total-5760: wa(850), fa(480), ru(850), &CSI Denoising (Lowpass filter, PCA), \\
& &(NTx, NRx, NSc) = (1, 3, 30) &si(590), pi(710), pu(760), wa(760), &Single-activity effective area \\
& &Such array is taken for all &bo(760) ; Trained on a 1 person's data, &determination \\
& &the 3 Transmit antennas &Tested on 6 persons' data & \\
& & & & \\
Damodaran et al. \cite{ubuntucsi} &1 &- &Total-1000 (200 each) &Denoising (DWT) \\
& & & & \\
Ashleibta et al. \cite{CSI5G} &1 &CSI array shape per sample &Total-1777 Samples. &Calculate only amplitudes from complex \\
& &(NTx, NRx, NSc) = (1, 1, 51) &Empty (117) &CSI data, then average the CSI \\
& & &1 person stand/sit/walk (140 each) & amplitudes of all subcarriers, \\
& & &2/3/4 person stand/sit (100 each) &Butterworth low-pass filtering of \\
& & &1 sit + 1 stand (100), 1 sit + 2 stand (120) &the average \\
& & &2 sit + 1 stand (100), 2 sit + 2 stand (100) & \\
& & &1 walk + 1 sit (100), 1 walk + 2 sit (120) & \\
& & &80:20 splitting for Train:Test & \\
& & & & \\
Shalaby et al. \cite{csiRev_cnnGRUattention} &1 &CSI array shape per sample &Total (8959): Li-1318, Fa-889, &Calculate only amplitudes from \\
& &(NTx, NRx, NSc) = (1, 3, 30) &Wa-2931, Ru-2408, Si-812, St-601 &complex CSI data \\
& & &Data from six persons. & \\
& & &Both the 70:30 train-test split and k-fold & \\
& & &cross-validation methods used. & \\
& & & & \\
Yang et al. \cite{autofi} &1 &CSI array shape &Gesture recognition training &The Atheros tool gives the only \\
& &(NRx, NSc, WiFi &samples = 5000, test samples = & amplitude of CSI. No preprocessing. \\
& &packets) = (3, 114, 500) &120 per category & \\
& & & & \\
Bocus et al. \cite{slfspcsi} &1 &CSI array shape per sample &Data from 6 participants. &Only CSI amplitude was used, denoising \\
& &(NTx, NRx, NSc) = (3, 3, 30) &80:20 train-test split &CSI amplitude using DWT, \\
& & & &segmentation into fixed duration, \\
& & & &dimensionality reduction using PCA \\
& & & & \\
Xu et al. \cite{contrst_csi} &2 &CSI array shape: (Time, NSc, NRx, NTx) &Office Room: 557 samples (6 participants) &- \\
& &Office Room Dataset: (T, 30, 3, 1) &Falldefi: 581 samples (3 participants) & \\
& &Falldefi Dataset: (10000, 30, 3, 1) &Signfi: 2760 samples (5 participants) & \\
& &Signfi Dataset: (200, 30, 3, 1) & & \\
& & & & \\
Al-qaness \cite{tailorcsi} &1 &CSI array shape per sample &Total 27000 samples &Amplitude and phase extraction, Denoising \\
& &(NTx, NRx, NSc) = (3, 3, 30) &10th fold cross-validation &(lowpass Butterworth filtering), PCA, \\
& & &used during training &Pattern Segmentation (envelop \\
& & & &extraction using Hilbert-Huang \\
& & & &transform to determine the width \\
& & & &of the dynamic time window) \\
& & & & \\
This Article &7 &Timestamp, Noise, AGC, RSSI at &Exp-1: data from 1 subject-pair. 120 trial files &segmentation into 1560 \\
& &3 Rx antennas, complex valued CSI. &(10 files for each of the steady-state followed &wifi packets \\
& &CSI array shape per wifi-packet: &by any other out of 12 activities). & \\
& &(NTx, NRx, NSc) = (2, 3, 30) &Exp-2: data from 10 subject-pair. 1200 trial files & \\
& & &Each file has around 1450-1650 wifi-packets/trial. & \\
& & &60:20:20 data-split for training-validation- & \\
& & &test set. 4-fold cross-validation used & \\
& & &during training. & \\
\bottomrule

\end{tabular}}
\end{adjustbox}

\label{tab: comptable3}
\end{sidewaystable}

\begin{sidewaystable}
\caption{A Comprehensive Survey and Comparative Analysis of Related Works to this Article (contd.)}
\centering
\begin{adjustbox}{width=\smallestside-1.3in}
{\begin{tabular}{lllllll}\toprule
\textbf{Article} &\textbf{Manually extracted} &\textbf{Proposed model} &\textbf{Performance metrics} &\textbf{Time} &\textbf{GUI} \\
\textbf{} &\textbf{features and final features} &\textbf{} &\textbf{} &\textbf{} &\textbf{software} \\
\textbf{} &\textbf{used for the classifier model} &\textbf{} &\textbf{} &\textbf{} &\textbf{developed?} \\\midrule
Wang et al. \cite{ieeelstm1} &Short Time Fourier Transform (STFT) on &CNN-LSTM sequential model &Acc: 0.78-0.83, Pre: 0.96-1.00, &Training time: 10 min &No \\
&both time and frequency domain signal & &Rec: 0.93-1.00, F1: 0.95-0.99 &Test time: 0.1 sec/activity & \\
&from only CSI amplitude and then spectrogram & &(For linear antenna setup) & & \\
&plotted from STFT result (image feature set as input). & & & & \\
& & & & & \\
Damodaran et al. \cite{ubuntucsi} &Only for SVM classifier: Dimensionality reduction &SVM model &Walk, Run: (pre, rec, F1) = (0.95, 0.98, 0.96) &- &No \\
&(PCA), feature extraction (PSD, frequency of center & &Sit, Stand: (pre, rec, F1) = (0.97, 0.95, 0.96) & & \\
&of energy, Haart wavelet analysis) &LSTM model &Walk, Run: (pre, rec, F1) = (0.90, 0.88, 0.89) & & \\
& & &Sit, Stand: (pre, rec, F1) = (0.88, 0.90, 0.89) & & \\
& & &Empty: all metrics = 1 & & \\
Ashleibta et al. \cite{CSI5G} &3-level DWT of the averaged data &1D CNN sequential model &Empty and all types of stand only (Acc: 0.9545) &- &No \\
& & &Empty and all types of sit only (Acc: 0.9134) & & \\
& & &Empty and all types of sit and stand (Acc: 0.8764) & & \\
& & &All 16 classes (Acc: 0.8567) & & \\
& & & & & \\
Shalaby et al. \cite{csiRev_cnnGRUattention} &- &CNN-GRU &Acc: 0.9946, Pre: 0.9952, Rec: 0.9943, AUC: 0.9990, Loss: 0.0026 &- &No \\
& &CNN-GRU-Attention &Acc: 0.9905, Pre: 0.9914, Rec: 0.9901, AUC: 0.9977, Loss: 0.0103 & & \\
& &CNN-GRU-CNN &Acc: 0.9905, Pre: 0.9909, Rec: 0.9903, AUC: 0.9974, Loss: 0.0411 & & \\
& &CNN-LSTM-CNN &Acc-0.9899, Pre: 0.9903, Rec: 0.9896, AUC: 0.9970, Loss: 0.0340 & & \\
& & & & & \\
Yang et al. \cite{autofi} &- &CNN-based geometric &Gesture recognition: Acc: 0.8971 using 3-shot learning &Training Epochs: 300 (for GSS), 100 (for FSC) &No \\
& &self-supervised learning &Gait analysis (i.e.: Human identification): &Batch Size: 128 & \\
& &followed by few-shot learning &Acc: 0.8333 using 3-shot learning &Training Time: 20 min (GSS), $<1$ min (FSC) & \\
& & & &Test time per sample: 22 msec & \\
& & & & & \\
Bocus et al. \cite{slfspcsi} &Conversion to multi-view spectrogram &Self-supervised contrastive &Best macro F1 score: 0.68 from AlexNet &Training Epoch: 200 &No \\
&image using STFT &pretraining using Shallow, & &Batch Size: 64 & \\
& &AlexNet, VGG16, ResNet & &Normalized Temperature & \\
& & & &Cross-entropy contrastive & \\
& & & &loss calculation & \\
& & & & & \\
Xu et al. \cite{contrst_csi} &- &Mutual information enhanced &Office Room: Acc: 0.9464, F1: 0.9464 &Training Epoch: 1500 &No \\
& &dual-stream contrastive &Falldefi: Acc: 0.8017, F1: 0.80 &(early stopping strategy used) & \\
& &learning unsupervised &Signfi: Acc: 0.8744, F1: 0.8655 & & \\
& &framework (DualConFi-FT) & & & \\
& & & & & \\
Al-qaness \cite{tailorcsi} &12 features (mean, maximum value, standard &Random Forest Classifier &Pre: 0.90-0.97 &- &No \\
&deviation, percentiles, median & &Rec: 0.8835-1.00 & & \\
&absolute deviation, and entropy) & &F1: 0.8967-0.9848 & & \\
&for both the amplitude and phase & & & & \\
& & & & & \\
This Article &time-difference calculation, &Attention-BiGRU &Exp-1: 1 subject-pair &Training Epochs: 300 (early stopping strategy used) &Yes \\
&3D to 2D array conversion of CSI data, &functional structured &Acc: 0.937, Pre: 0.944, Rec: 0.937, F1: 0.935 &Batch Size: 12 & \\
&calculation of amplitude and phase &supervised time-series &Exp-2: 10 subject-pair &Learning rate tuning: ReduceLROnPlateau & \\
&of CSI data, creation of $ 1560 \times 366 $ &classification &Acc: 0.883, Pre: 0.887, Rec: 0.883, F1: 0.884 &Training Time: 1-1.5hrs/fold (Exp-1), 7-8hrs/fold (Exp-2) & \\
&shaped 2D feature set (from time-diff, &deep-learning model & &Test time per trial file: 35 seconds (using the & \\
&noise, AGC, RSSI, amplitude and & & &learned parameters from all the 4-fold training). & \\
&phase of CSI) & & &09 seconds (using the learned & \\
& & & & parameters from a single-fold training) & \\

\bottomrule

\end{tabular}}
\end{adjustbox}

\label{tab: comptable4}
\end{sidewaystable}

In contrast to previous literature, which focused on single activities conducted by individuals over a fixed time period, often with less than ten target classes, this article presents a novel approach to mutual human interaction recognition utilizing a HAR dataset that captures WiFi channel-state variations during mutual-concurrent human interactions of varying lengths. The proposed methodology demonstrates significant improvements in identifying and classifying similar interactions, such as approaching vs departing, hugging vs pushing vs pointing vs punching, kicking with left vs right leg, and pointing/punching with left vs right hand. These results highlight the effectiveness of the proposed deep learning solution, which incorporates self-attention mechanism, position embedding, and BiGRU to achieve superior classification performance.

The proposed deep learning solution in this study not only achieves accurate classification but also offers a novel interactive graphical user interface (GUI) software. Unlike existing literature, this software enables users to visualize the classification results in a time-series representation and save them for further analysis. The GUI software is highly flexible, allowing users to classify a single trial file or multiple files stored in a folder. Furthermore, it takes only 9 seconds to preprocess, classify, save, and plot the data on the interactive plotly dashboard using the learned parameters from a single-fold training. Additionally, it can improve classification accuracy by automatically incorporating (if *.h5 file found in the `cloverleaf' folder of the software) unlimited trained parameter files obtained from all k-fold training. The latter approach is seen to take 35 seconds for the trained parameter files obtained through 4-fold training.

The recent literature has mainly focused on using the SIMO/SISO CSI amplitude value as a feature from their datasets. However, this article took seven features into account, including the phase angle of MIMO CSI data, CSI amplitude value, WiFi packet arrival time-difference, noise, AGC, and RSSI at three receive antennas, obtained from the HAR dataset \cite{dataset}. Unlike previous studies, this research incorporated a large volume of data and worked on raw data without any denoising  (using lowpass Butterworth filtering or Discrete Wavelet Transform i.e. DWT), dimensionality reduction  (using Principle Component Analysis, i.e. PCA) or feature extraction via signal processing (using STFT, Haart wavelet analysis, spectrogram, Power Spectral Density i.e. PSD, etc.) techniques. Due to memory constraints provided by Google Colaboratory's free plan, the study used data from only 10 subject pairs, while the original dataset contained data from 40 subject pairs. The authors of this study believe that not using a dimensionality reduction technique might have resulted in a longer model training time. However, they also believe that this approach preserved crucial information and accurately classified WiFi packets during the concurrent activities' transition period.

The proposed deep learning solution presented in this article demonstrated significant and satisfactory performance in experiments 1 and 2 despite the challenges of a LOS-faded MIMO WiFi-NIC radio link, mutuality, concurrency, inequal time-length, inter-activity variability, and memory constraints. However, experiment 3 revealed the Domain-Shift Problem \cite{domain_shift}, which could be addressed in future research through the use of Domain Adaptation \cite{domain_adaptation1, domain_adaptation2} techniques and Self-supervised contrastive training followed by few-shot learning strategies (\cite{contrst_csi}, \cite{autofi}, \cite{slfspcsi}), as well as more sophisticated signal pre-processing techniques and inclusion of CSI data for all 51 subcarriers.

\section{Future Scope}
\label{sec:scopes}
This research work developed the mutual interaction recognition model based on the dataset collected in a laboratory setting. Public deployment of such a recognition model still requires to overcome enormous challenges to achieve the pinnacle. One such obstacle is to explore and overcome the interclass pattern similarity problem due to different interactions having almost similar gesture-style. Another one is intraclass pattern variation due to dissimilar height-weight-clothing-gesture style of different subject pairs involved in mutual interaction. Sensitive environmental conditions such as weather, noise, inter-device interference, occlusions due to indefinite indoor objects, transmitter-receiver separation and location of the interaction occurrences within the room also contribute a lot to the intraclass pattern variation in real-world scenario. 

To reach the goal, more efficient pattern diversities might be discovered in future via extraction of additional informative features through incorporation of advanced signal processing techniques. Simultaneous inclusion of model-based approaches along with the deep learning approaches and inheriting domain adaptation techniques can be a promising remedy to the changes in the data distribution problem. Discovering alternative to network interface card for CSI data acquisition containing information of all the 51 subcarriers and calibrating the model with a large volume of training instances characterized by wide-ranging environmental and fabricated factors are still a few paramount areas of investigation that need to be unfolded. Another promising successor of this proposed article can be indoor human-pair occupancy counting. To get the ultimate implications of this study, future research work could also address concurrent recognition of spontaneous (another interaction whilst in the middle of a particular interaction, e.g. sudden kick while departing) and multiple mutual interactions performed by any number of subject-pairs present in the room or by more than two individuals in a group.

In this study, the Python Keras framework has been utilized which provides a built-in deep neural network optimizer library, specifically the Adam optimizer, which has been shown to outperform \cite{adambest} other available optimizer functions. The modified Stochastic Gradient Descent based optimizer algorithm, i.e., Adam optimizer, was used to optimize hyper-parameters, weights, number of layers, number of neurons, learning rate, etc. for the proposed Attention-BiGRU deep learning architecture. However, traditional iterative forward and back propagation based gradient descent deep learning optimization algorithms can be computationally burdensome for complex and large datasets. These algorithms require the loss (also known as, cost) functions to be continuous in order to compute the gradients and often struggle to reach the global minimum and get stuck at the local minimum. Additionally, they require a long training time for the optimization algorithm. In experiment-1 of this study, which consisted of data from a single subject-pair, the training time required per fold out of the four-fold training sets was approximately 1-1.5 hours. In experiment-2, which consisted of data from ten subject pairs, the training time per-fold out of the four-fold training sets reached approximately 7-8 hours. Recent advancements in deep-learning model optimization algorithms have aimed to address these problems associated with traditional iterative back-propagation based gradient descent optimization algorithms. Although there may be some drawbacks and challenges in applying custom-built algorithms to deep-learning model optimization, biologically-inspired methods known as Meta-Heuristics (MH) optimization have shown promise in optimizing deep-learning architectures. Kaveh et al. \cite{MHopt1}, Elaziz et al. \cite{MHopt2}, and Khan et al. \cite{MHopt3} have extensively reviewed the latest developments in the use of MH algorithms for optimizing deep-learning architectures in order to find out the best combination of weights, no. of layers and nodes, learning parameters, other hyper-parameters, and activation function. In general, there are four categories of MH algorithms, namely Evolutionary Algorithms, Swarm Intelligence, Natural Phenomena, and Human Inspiration, each of which comprises several subtypes. These different types and sub-types of MH algorithms have been extensively discussed in the aforementioned review articles. Such MH optimization algorithms have shown promising results in tackling complex optimization problems with the ability to escape local minima and reach the global minima in a shorter training period through using exploration and exploitation techniques. These improve the model accuracy and speed up the execution time, even when dealing with discontinuous loss (also known as, cost) functions. While the present study utilized only the Adam optimizer, future investigations should consider exploring custom-built MH optimizers as discussed in the research articles by Kaveh et al. \cite{MHopt1}, Elaziz et al. \cite{MHopt2}, Khan et al. \cite{MHopt3} as well as the hybrid multi-MH technique such as Invasive Weed Optimization integrated with Differential Evolutionary Model as proposed by Movassagh et al. \cite{MHopt4}. In addition, incorporation of Dynamic Programming (DP) \cite{DP1, DP2, DP3} with the proposed Attention-BiGRU deep learning architecture may enhance optimization speed and improve model accuracy. Because, DP divides the entire task into multiple sub-tasks and applies a bottom-up solution approach to solve each sub-task while storing the result in memory for later use. By doing so, DP finds solutions with fewer computations, saving time.

In the context of developing an Internet of Things (IoT)-based smart city, it is crucial to address three key aspects - heterogeneity, privacy preservation of generated data, and provision of high-level services - due to the large number of devices developed by multiple manufacturers, as emphasized by Gheisari et al. \cite{obpp}. The proposed software solution in the current study for the human to human mutual interaction recognition ensures privacy of the indoor residents as neither being intrusive nor obtrusive, as discussed in Section-\ref{sec:Introduction}. However, additional measures must be taken in future to protect data from third-party breaches. In this study, the experimental analyses performed using a dataset \cite{dataset} for which a publicly available CSI tool \cite{csi_tool} was used to collect WiFi channel data from a three-antenna Intel 5300 NIC WiFi receiver mounted on a personal computer, while a two-antenna WiFi access point (model: Sagemcom-2704) developed by Sagemcom Broadband SAS was used as the WiFi transmitter. The device configuration and experimental setup are shown in Figure-\ref{fig: MIMO}. If data were collected from a device developed by a different vendor or data were collected from the same Sagemcom device with different device settings, the heterogeneity problem described by Gheisari et al. \cite{obpp} would arise, and the proposed software solution for human activity recognition might not be consistent. Additionally, since CSI extraction using a CSI extraction tool must currently be done manually, steps should be taken to fully automate the system, including csi extraction and feeding the extracted data to the proposed software framework, through the development of smart IoT devices. Moreover, more data should be trained to provide high-level services for the device. Finally, an `Ontology-Based Privacy-Preserving (OBPP)’ framework proposed by Gheisari et al. \cite{obpp} may be employed to address the three aforementioned challenges in the development of integrated smart IoT devices for WLAN based human-to-human mutual interaction recognition.

\section{Conclusion}
\label{sec:conclusion}
This paper presents a WiFi-based deep learning solution for mutual concurrent human-to-human interaction recognition in an indoor environment through a MIMO communication link between a WiFi router and an Intel 5300 NIC mounted in a computer. To improve the learning of the channel state variations due to mutual human activities, the phase angle of the CSI arrays has been included in the analysis along with the CSI amplitude. WiFi packet reception time interval, accumulated noise, received signal strength indicator, and automatic gain control parameters have also been considered as feature sets in order to attain robust information about the signal changes and received signal quality during performance of mutual human activities. We overcome the complexities set by concurrent, very similar mutual activities performed during unequal time lengths by employing Positional Embedding to weight data by temporal position, and then employing Multi-Head Self-Attention to score data samples based on their self-interrelation in multiple contexts. We use Bi-directional Gated Recurrent Neural Network to learn time-series features from CSI and received signal quality data, reducing computational complexity with fewer gates and faster computation. The proposed Attention-BiGRU deep learning functional architecture is set to classify thirteen mutual activities in three experiments. The experimental analysis reveals that the proposed solution performs the best if it is trained and tested on a single subject pair. The model still achieves a considerable time-series classification performance if it is trained and tested on upto ten subject pairs, indicating that the suggested model can endure a subtle amount of inter-activity variability. In fact, it is possible to classify, save, and observe the temporal extent of mutual interactions for a specific subject pair using the developed executable GUI software. However, the proposed solution is yet to be capable enough to be employed on untrained subject-pairs. The paper discusses ways to improve the mutual interaction recognition architecture and notes that the method preserves privacy as it doesn't capture visual data. The cognitive ability of such technology possesses significant potential for indoor monitoring, surveillance, health monitoring, and assisted living.

\section{Supplementary Materials}
\label{sec:supp_link}

\begin{itemize}
    \item[-] All Learning Curves and Confusion Matrices: supplementary\_materials/sup1\_All\_Performance\_Figures.pdf
    \item[-] GUI Application Demonstration Video: supplementary\_materials/sup2\_H2HI\_WiFi\_GUI\_App\_Demo\_Video.mp4
    \item[-] GUI Application GitHub Repository: Link will be added to the published version of the article.

\end{itemize}

\section{Author Statements}
\label{sec:auth_contr}
\noindent\textit{Md. Mohi Uddin Khan:} Idea generation, Methodology, Formal analysis, Research validation, Visualization, Software, Writing - Original Draft, Editing, and Quality control.

\vspace{4pt}

\noindent\textit{Abdullah Bin Shams and Md. Mohsin Sarker Raihan:} Idea generation, Supervision, Research validation, Writing - Quality control.

\vspace{4pt}

\noindent Finally all authors reviewed and discussed on the article, provided critical feedback and contributed to the final manuscript.

\section{Research Funding}
\label{sec:funding}
This research did not receive any specific grant from funding agencies in the public, commercial, or
not-for-profit sectors.

\section{Declaration of Competing Interest}
\label{sec:interest}
The authors declare that they have no known competing financial interests or personal relationships that could have appeared to influence the work reported in this article.

\section{Data Availability Statement}
\label{sec:DAS}
The dataset \cite{dataset} analyzed during the current study is available in the Mendeley Data Repository, \href{https://data.mendeley.com/datasets/3dhn4xnjxw/1}{https://data.mendeley.com/datasets/3dhn4xnjxw/1}.

\clearpage
\begin{appendices}
\section{MIMO-OFDM}
\label{appendix:mimo_ofdm}
\begin{figure*}[ht]
\centering 
\resizebox{\textwidth}{!}{\includegraphics{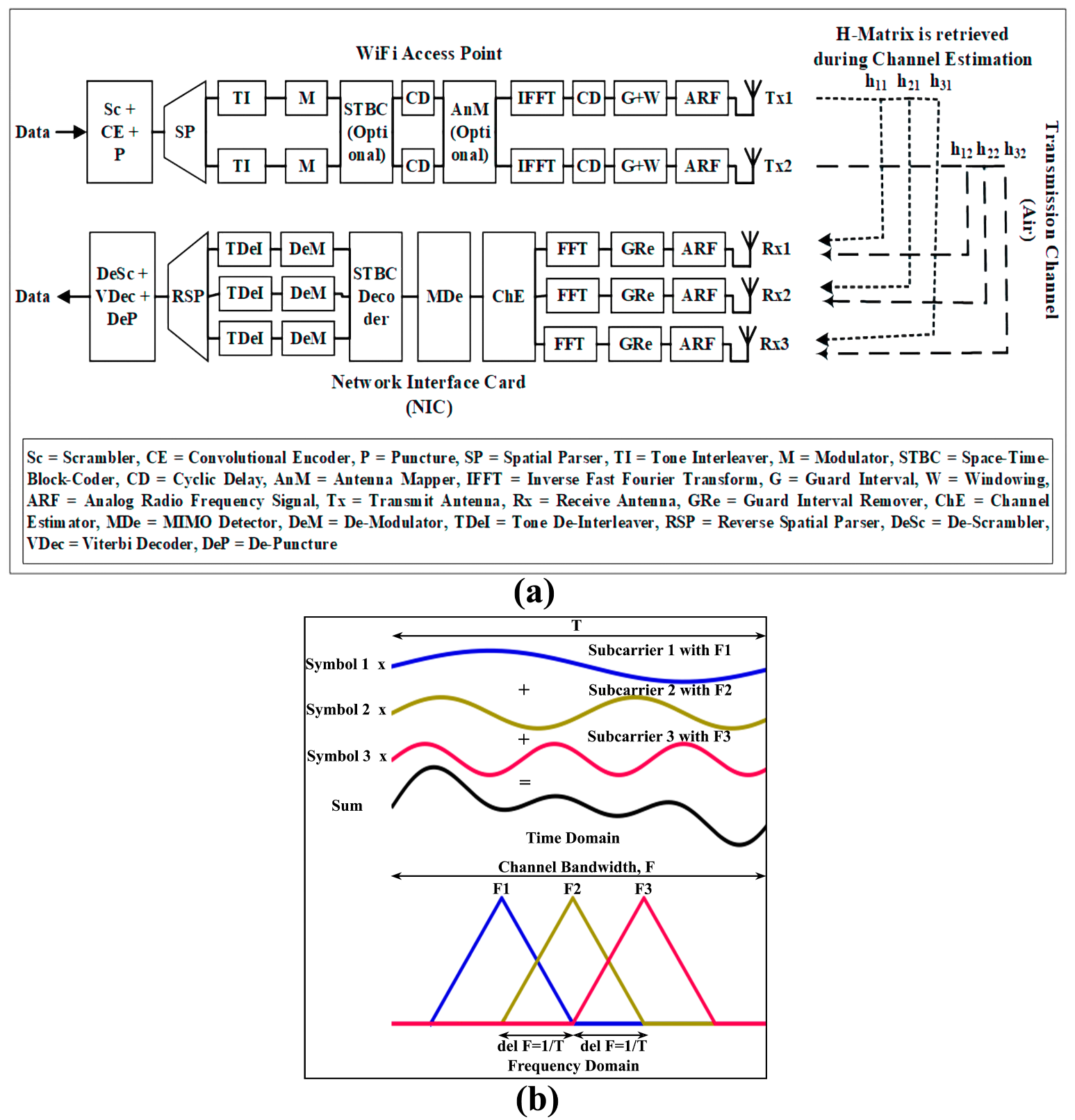}}
\caption{{(a) Generalized block diagram of 2x3 MIMO-OFDM transmission and reception structures according to IEEE 802.11n standard (b) OFDM Intuition using three subcarriers}}
\label{fig: MIMO_OFDM}
\end{figure*}

With the advent of high-speed wireless local area network (WLAN) communication techniques, a multi-antenna system called Multiple Input Multiple Output (MIMO) (Figure-\ref{fig: MIMO_OFDM}(a)) has increased the throughput, spectral efficiency, quality-of-service (QoS) as well as transmission range by means of signal fading reduction after utilizing the Orthogonal Frequency Division Multiplexing (OFDM) (Figure-\ref{fig: MIMO_OFDM}(b)) transmission scheme along with IEEE 802.11n amendment compared to IEEE 802.11a/b/g standards ensuring the use of 2.4 GHz carrier frequency band and 20 MHz (optional 40 MHz) channel bandwidth which brought down the production cost and throughput maximized to 600 Mbps. \cite{IEEE80211n}

Previously, for Frequency Division Multiplexing (FDM), the overall channel bandwidth had to be equally divided for N subcarriers with addition of guard interval between two subcarriers. But, the invention of calculation-efficient Discrete Fast Fourier Transform (DFFT) using matrix-algebra, affordable high-speed VLSI circuit design have geared up the development of OFDM modulation scheme. 

Analog baseband signal is converted to digital (with A/D converter) followed by symbol creation via constellation mapper. The overall channel bandwidth is divided into N (51 in this experiment, note that Intel NIC can report only 30) non-overlapping subcarrier bandwidth in such a way that subcarrier-spacing $\left( \Delta f=\frac{1}{T} \right)$ is minimum and respective signals are orthogonal to each other. Then the symbols are modulated with particular subcarrier frequency. After addition of cyclic prefix for taking care of Inter-Symbol-Interference (ISI) and other processing, the frequency domain signal is converted to time domain analog signal using Inverse-FFT for radio wave transmission generated by transmitter oscillator. Orthogonal property of the subcarriers prevents the Inter-Carrier-Interference. Symbol rate for each subcarrier needs to be less so as to reduce aliasing and multipath propagation effects but that is compensated by increased number of orthogonal subcarriers which all in all increased the net transmission rate (throughput) via efficient use of available channel spectrum. \cite{shankar, phdthesis, hanzo}
\end{appendices}

\clearpage
\bibliography{references}

\begin{thebibliography}{94}
\providecommand{\natexlab}[1]{#1}
\providecommand{\url}[1]{{#1}}
\providecommand{\urlprefix}{URL }
\providecommand{\doi}[1]{\url{https://doi.org/#1}}
\providecommand{\eprint}[2][]{\url{#2}}
 \bibcommenthead

\bibitem[{Gowda et~al(2023)Gowda, Shetty, Darshini, and Rajani}]{comboreview7}
Gowda SG, Shetty SM, Darshini MS, et~al (2023) Analysis of human activity
  detection using machine learning approaches. SN Computer Science 4(2).
  \doi{10.1007/s42979-022-01550-x}

\bibitem[{Saleem et~al(2022)Saleem, Bajwa, and Raza}]{comboreview6}
Saleem G, Bajwa UI, Raza RH (2022) Toward human activity recognition: A survey.
  Neural Computing and Applications 35(5):4145–4182.
  \doi{10.1007/s00521-022-07937-4}

\bibitem[{Kulsoom et~al(2022)Kulsoom, Narejo, Mehmood, Chaudhry, Butt, and
  Bashir}]{comboreview5}
Kulsoom F, Narejo S, Mehmood Z, et~al (2022) A review of machine learning-based
  human activity recognition for diverse applications. Neural Computing and
  Applications 34(21):18289–18324. \doi{10.1007/s00521-022-07665-9}

\bibitem[{Gupta et~al(2022)Gupta, Gupta, Pathak, Jain, Rashidi, and
  Suri}]{comboreview4gupta}
Gupta N, Gupta SK, Pathak RK, et~al (2022) Human activity recognition in
  artificial intelligence framework: A narrative review. Artificial
  Intelligence Review 55(6):4755–4808. \doi{10.1007/s10462-021-10116-x}

\bibitem[{Islam et~al(2022)Islam, Nooruddin, Karray, and
  Muhammad}]{comboreview4Islam}
Islam MM, Nooruddin S, Karray F, et~al (2022) Human activity recognition using
  tools of convolutional neural networks: A state of the art review, data sets,
  challenges, and future prospects. Computers in Biology and Medicine
  149:106060. \doi{10.1016/j.compbiomed.2022.106060}

\bibitem[{Beddiar et~al(2020)Beddiar, Nini, Sabokrou, and Hadid}]{comboreview3}
Beddiar DR, Nini B, Sabokrou M, et~al (2020) Vision-based human activity
  recognition: A survey. Multimedia Tools and Applications
  79(41-42):30509–30555. \doi{10.1007/s11042-020-09004-3}

\bibitem[{Zhang et~al(2017)Zhang, Wei, Nie, Huang, Wang, and Li}]{comboreview2}
Zhang S, Wei Z, Nie J, et~al (2017) A review on human activity recognition
  using vision-based method. Journal of Healthcare Engineering 2017:1–31.
  \doi{10.1155/2017/3090343}

\bibitem[{Ranasinghe et~al(2016)Ranasinghe, Al~Machot, and Mayr}]{comboreview1}
Ranasinghe S, Al~Machot F, Mayr HC (2016) A review on applications of activity
  recognition systems with regard to performance and evaluation. International
  Journal of Distributed Sensor Networks 12(8). \doi{10.1177/1550147716665520}

\bibitem[{Golestani and Moghaddam(2020)}]{MIsensor}
Golestani N, Moghaddam M (2020) Human activity recognition using magnetic
  induction-based motion signals and deep recurrent neural networks. Nature
  Communications 11(1). \doi{10.1038/s41467-020-15086-2}

\bibitem[{Li et~al(2020)Li, Li, Tyson, and Xie}]{camprivacy1}
Li J, Li Z, Tyson G, et~al (2020) Your privilege gives your privacy away: An
  analysis of a home security camera service. In: IEEE INFOCOM 2020 - IEEE
  Conference on Computer Communications, pp 387--396,
  \doi{10.1109/INFOCOM41043.2020.9155516}

\bibitem[{Li et~al(2016)Li, He, Sun, Cheng, and Yu}]{camprivacy2}
Li H, He Y, Sun L, et~al (2016) Side-channel information leakage of encrypted
  video stream in video surveillance systems. In: IEEE INFOCOM 2016 - The 35th
  Annual IEEE International Conference on Computer Communications, pp 1--9,
  \doi{10.1109/INFOCOM.2016.7524621}

\bibitem[{Xu et~al(2023)Xu, Wang, Zhang, Zhu, and Zheng}]{contrst_csi}
Xu K, Wang J, Zhang L, et~al (2023) Dual-stream contrastive learning for
  channel state information based human activity recognition. IEEE Journal of
  Biomedical and Health Informatics 27:329--338.
  \doi{10.1109/JBHI.2022.3219640}

\bibitem[{McCaldin et~al(2016)McCaldin, Wang, Schreier, Lovell, Marschollek,
  Redmond, and Schukat}]{sensorprivacy}
McCaldin D, Wang K, Schreier G, et~al (2016) Unintended consequences of
  wearable sensor use in healthcare. Yearbook of Medical Informatics
  25(01):73–86. \doi{10.15265/iy-2016-025}

\bibitem[{Ding et~al(2021)Ding, Guo, and Wang}]{radar1}
Ding W, Guo X, Wang G (2021) Radar-based human activity recognition using
  hybrid neural network model with multidomain fusion. IEEE Transactions on
  Aerospace and Electronic Systems 57(5):2889--2898.
  \doi{10.1109/TAES.2021.3068436}

\bibitem[{Noori et~al(2021)Noori, Uddin, and Torresen}]{radar2}
Noori FM, Uddin MZ, Torresen J (2021) Ultra-wideband radar-based activity
  recognition using deep learning. IEEE Access 9:138132--138143.
  \doi{10.1109/ACCESS.2021.3117667}

\bibitem[{Ghosh et~al(2017)Ghosh, Sanyal, Chakraborty, Sharma, Saha, Nandi, and
  Saha}]{ultrasonic}
Ghosh A, Sanyal A, Chakraborty A, et~al (2017) On automatizing recognition of
  multiple human activities using ultrasonic sensor grid. In: 2017 9th
  International Conference on Communication Systems and Networks (COMSNETS), pp
  488--491, \doi{10.1109/COMSNETS.2017.7945440}

\bibitem[{Damodaran and Schäfer(2019)}]{csibetter}
Damodaran N, Schäfer J (2019) Device free human activity recognition using
  wifi channel state information. In: 2019 IEEE SmartWorld, Ubiquitous
  Intelligence Computing, Advanced Trusted Computing, Scalable Computing
  Communications, Cloud Big Data Computing, Internet of People and Smart City
  Innovation (SmartWorld/SCALCOM/UIC/ATC/CBDCom/IOP/SCI), pp 1069--1074,
  \doi{10.1109/SmartWorld-UIC-ATC-SCALCOM-IOP-SCI.2019.00205}

\bibitem[{Yang et~al(2022)Yang, Xu, Cao, Zou, and Xie}]{wifi_attention_better}
Yang J, Xu Y, Cao H, et~al (2022) Deep learning and transfer learning for
  device-free human activity recognition: A survey. Journal of Automation and
  Intelligence 1(1):100007. \doi{10.1016/j.jai.2022.100007}

\bibitem[{Liu et~al(2022)Liu, Ramli, Zhang, Henricson, and
  Liu}]{concurrent_composite}
Liu R, Ramli AA, Zhang H, et~al (2022) An overview of human activity
  recognition using wearable sensors: Healthcare and artificial intelligence.
  Internet of Things – ICIOT 2021 p 1–14. \doi{10.1007/978-3-030-96068-1_1}

\bibitem[{Ye et~al(2020)Ye, Zhong, Qu, and Zhang}]{videoh2hi1}
Ye Q, Zhong H, Qu C, et~al (2020) Human interaction recognition based on
  whole-individual detection. Sensors 20(8). \doi{10.3390/s20082346},
  \urlprefix\url{https://www.mdpi.com/1424-8220/20/8/2346}

\bibitem[{Ahmed et~al(2019)Ahmed, Kim, Kim, Bashar, and Rhee}]{videoh2hi2}
Ahmed MU, Kim YH, Kim JW, et~al (2019) Two person interaction recognition based
  on effective hybrid learning. KSII Transactions on Internet and Information
  Systems 13(2). \doi{10.3837/tiis.2019.02.015}

\bibitem[{Stergiou and Poppe(2019)}]{videoh2hi3}
Stergiou A, Poppe R (2019) Analyzing human–human interactions: A survey.
  Computer Vision and Image Understanding 188:102799.
  \doi{https://doi.org/10.1016/j.cviu.2019.102799},
  \urlprefix\url{https://www.sciencedirect.com/science/article/pii/S1077314219301158}

\bibitem[{Liu et~al(2020)Liu, Yang, Yang, Hou, Hu, and Jiang}]{radarh2hi}
Liu H, Yang R, Yang Y, et~al (2020) Human–human interaction recognition based
  on ultra-wideband radar. Signal, Image and Video Processing
  14(6):1181–1188. \doi{10.1007/s11760-020-01658-8}

\bibitem[{Finkelhor and Ormrod(2001)}]{babysit}
Finkelhor D, Ormrod R (2001) Crimes against children by babysitters. Juvenile
  Justice Bulletin NCJ198102:1–7. \doi{10.1037/e317992004-001},
  \urlprefix\url{https://scholars.unh.edu/ccrc/9/}

\bibitem[{Gerard et~al(2018)Gerard, McGrath, Colvin, and
  McFarlane}]{carecenter}
Gerard A, McGrath A, Colvin E, et~al (2018) ‘i’m not getting out of bed!’
  the criminalisation of young people in residential care. Australian \& New
  Zealand Journal of Criminology 52(1):76–93. \doi{10.1177/0004865818778739}

\bibitem[{Hemphill et~al(2009)Hemphill, Smith, Toumbourou, Herrenkohl,
  Catalano, McMorris, and Romanuik}]{youthviolence}
Hemphill SA, Smith R, Toumbourou JW, et~al (2009) Modifiable determinants of
  youth violence in australia and the united states: A longitudinal study.
  Australian \& New Zealand Journal of Criminology 42(3):289–309.
  \doi{10.1375/acri.42.3.289}

\bibitem[{Crist\'{o}bal and Boettcher(2018)}]{hazing1}
Crist\'{o}bal S, Boettcher ML (2018) Critical perspectives on hazing in
  colleges and universities: A guide to disrupting hazing culture, 1st edn.
  Routledge, \urlprefix\url{https://doi.org/10.4324/9781315177311}

\bibitem[{Allan and Madden(2012)}]{hazing2}
Allan EJ, Madden M (2012) The nature and extent of college student hazing.
  International Journal of Adolescent Medicine and Health 24(1).
  \doi{10.1515/ijamh.2012.012}

\bibitem[{Campo et~al(2005)Campo, Poulos, and Sipple}]{hazing3}
Campo S, Poulos G, Sipple JW (2005) Prevalence and profiling: Hazing among
  college students and points of intervention. American Journal of Health
  Behavior 29(2):137–149. \doi{10.5993/ajhb.29.2.5}

\bibitem[{Kozicki et~al(1991)Kozicki, Hoenig, and Robinson}]{cleanroom1}
Kozicki M, Hoenig S, Robinson P (1991) Personnel and contamination, Springer, p
  211–251. \doi{10.1007/978-94-011-7950-8_11}

\bibitem[{Alavi-Moghadam et~al(2020)Alavi-Moghadam, Sarvari, Goodarzi, and
  Aghayan}]{cleanroom2}
Alavi-Moghadam S, Sarvari M, Goodarzi P, et~al (2020) The Importance of
  Cleanroom Facility in Manufacturing Biomedical Products, Springer, p 69–79.
  \doi{10.1007/978-3-030-35626-2_7}

\bibitem[{Alazrai et~al(2020)Alazrai, Awad, Alsaify, Hababeh, and
  Daoud}]{dataset}
Alazrai R, Awad A, Alsaify B, et~al (2020) A dataset for wi-fi-based
  human-to-human interaction recognition. Data in Brief 31:105668.
  \doi{10.1016/j.dib.2020.105668}

\bibitem[{Willman(2020)}]{pyqt5}
Willman J (2020) Overview of pyqt5. Modern PyQt p 1–42.
  \doi{10.1007/978-1-4842-6603-8_1}

\bibitem[{Thariq~Ahmed et~al(2020)Thariq~Ahmed, Ahmad, and C.V.}]{csi_review}
Thariq~Ahmed HF, Ahmad H, C.V. A (2020) Device free human gesture recognition
  using wi-fi csi: A survey. Engineering Applications of Artificial
  Intelligence 87:103281. \doi{10.1016/j.engappai.2019.103281}

\bibitem[{Ashleibta et~al(2021)Ashleibta, Taha, Khan, Taylor, Tahir, Zoha,
  Abbasi, and Imran}]{CSI5G}
Ashleibta AM, Taha A, Khan MA, et~al (2021) 5g-enabled contactless multi-user
  presence and activity detection for independent assisted living. Scientific
  Reports 11(1). \doi{10.1038/s41598-021-96689-7}

\bibitem[{Wang et~al(2021)Wang, Huang, Zhang, Dou, Guo, and
  Chen}]{modelbasedcsi_review}
Wang Z, Huang Z, Zhang C, et~al (2021) Csi-based human sensing using
  model-based approaches: A survey. Journal of Computational Design and
  Engineering 8(2):510–523. \doi{10.1093/jcde/qwab003}

\bibitem[{Wang et~al(2019)Wang, Gong, and Liu}]{ieeelstm1}
Wang F, Gong W, Liu J (2019) On spatial diversity in wifi-based human activity
  recognition: A deep learning-based approach. IEEE Internet of Things Journal
  6(2):2035--2047. \doi{10.1109/JIOT.2018.2871445}

\bibitem[{Wang et~al(2020)Wang, Gong, Liu, and Wu}]{ieeelstm2}
Wang F, Gong W, Liu J, et~al (2020) Channel selective activity recognition with
  wifi: A deep learning approach exploring wideband information. IEEE
  Transactions on Network Science and Engineering 7(1):181--192.
  \doi{10.1109/TNSE.2018.2825144}

\bibitem[{Damodaran et~al(2020)Damodaran, Haruni, Kokhkharova, and
  Schäfer}]{ubuntucsi}
Damodaran N, Haruni E, Kokhkharova M, et~al (2020) Device free human activity
  and fall recognition using wifi channel state information (csi). CCF
  Transactions on Pervasive Computing and Interaction 2(1):1–17.
  \doi{10.1007/s42486-020-00027-1}

\bibitem[{Halperin et~al(2011)Halperin, Hu, Sheth, and Wetherall}]{csi_tool}
Halperin D, Hu W, Sheth A, et~al (2011) Tool release: Gathering 802.11n traces
  with channel state information. ACM SIGCOMM Computer Communication Review
  41(1):53–53. \doi{10.1145/1925861.1925870}

\bibitem[{Ettus and Braun(2015)}]{usrp1}
Ettus M, Braun M (2015) The universal software radio peripheral (usrp) family
  of low-cost sdrs. Opportunistic Spectrum Sharing and White Space Access p pp.
  3–23. \doi{10.1002/9781119057246.ch1}

\bibitem[{Serkin and Vazhenin(2013)}]{usrp2}
Serkin F, Vazhenin N (2013) Usrp platform for communication systems research.
  In: 2013 15th International Conference on Transparent Optical Networks
  (ICTON), pp 1--4, \doi{10.1109/ICTON.2013.6602738}

\bibitem[{Shalaby et~al(2022)Shalaby, ElShennawy, and
  Sarhan}]{csiRev_cnnGRUattention}
Shalaby E, ElShennawy N, Sarhan A (2022) Utilizing deep learning models in
  csi-based human activity recognition. Neural Computing and Applications
  34(8):5993–6010. \doi{10.1007/s00521-021-06787-w}

\bibitem[{Yang et~al(2022)Yang, Chen, Zou, Wang, and Xie}]{autofi}
Yang J, Chen X, Zou H, et~al (2022) Autofi: Towards automatic wifi human
  sensing via geometric self-supervised learning. IEEE Internet of Things
  Journal pp 1--1. \doi{10.1109/JIOT.2022.3228820}

\bibitem[{Bocus et~al(2022)Bocus, Lau, McConville, Piechocki, and
  Santos-Rodriguez}]{slfspcsi}
Bocus MJ, Lau HS, McConville R, et~al (2022) Self-supervised wifi-based
  activity recognition. IEEE, pp 552--557,
  \doi{10.1109/GCWkshps56602.2022.10008537}

\bibitem[{Al-qaness(2019)}]{tailorcsi}
Al-qaness MAA (2019) Device-free human micro-activity recognition method using
  wifi signals. Geo-spatial Information Science 22:128--137.
  \doi{10.1080/10095020.2019.1612600}

\bibitem[{Li et~al(2019)Li, He, Li, and Jing}]{radarBiGRU}
Li C, He Y, Li X, et~al (2019) Bigru network for human activity recognition in
  high resolution range profile. In: 2019 International Radar Conference
  (RADAR), pp 1--5, \doi{10.1109/RADAR41533.2019.171259}

\bibitem[{Liu et~al(2020)Liu, You, Wu, Li, Li, Zhang, and Ge}]{grutrend1}
Liu X, You J, Wu Y, et~al (2020) Attention-based bidirectional gru networks for
  efficient https traffic classification. Information Sciences 541:297--315.
  \doi{https://doi.org/10.1016/j.ins.2020.05.035},
  \urlprefix\url{https://www.sciencedirect.com/science/article/pii/S002002552030445X}

\bibitem[{Chen et~al(2019)Chen, Jiang, and Zhang}]{grutrend2}
Chen JX, Jiang DM, Zhang YN (2019) A hierarchical bidirectional gru model with
  attention for eeg-based emotion classification. IEEE Access 7:118530--118540.
  \doi{10.1109/ACCESS.2019.2936817}

\bibitem[{Sheng(2020)}]{grutrend3}
Sheng L (2020) Application of attention-based gru combined with cnn
  classification on p300 signals. In: 2020 5th International Conference on
  Smart Grid and Electrical Automation (ICSGEA), pp 182--185,
  \doi{10.1109/ICSGEA51094.2020.00046}

\bibitem[{Paul and Ogunfunmi(2008)}]{IEEE80211n}
Paul T, Ogunfunmi T (2008) Wireless lan comes of age: Understanding the ieee
  802.11n amendment. IEEE Circuits and Systems Magazine 8(1):28–54.
  \doi{10.1109/mcas.2008.915504}

\bibitem[{Rappaport(2002)}]{rappaport}
Rappaport TS (2002) Wireless Communications: Principles and practice, Prentice
  Hall PTR, chap 3, 4, p 69–192

\bibitem[{Wong(2012)}]{daniel}
Wong KD (2012) Fundamentals of Wireless Communication Engineering Technologies,
  John Wiley \& Sons, chap 5: Propagation, p 125–154

\bibitem[{Stein(1998)}]{IndoorWLAN}
Stein JC (1998) Indoor radio wlan performance part ii : Range performance in a
  dense office environment. In: Intersil Corporation, 2401 Palm Bay, Florida
  32905

\bibitem[{SHANKAR(2017)}]{shankar}
SHANKAR PM (2017) Fading and shadowing in wireless systems, 2nd edn., Springer
  International Publishing AG, chap 3.8 Orthogonal Frequency Division
  Multiplexing, 4.3.1 Rayleigh Fading, 4.3.2 Rician Fading, pp 263, 303–267,
  312

\bibitem[{Weisstein and W.(2021)}]{besselfn1}
Weisstein, W. E (2021) Modified bessel function of the first kind.
  \urlprefix\url{https://mathworld.wolfram.com/ModifiedBesselFunctionoftheFirstKind.html}

\bibitem[{Abramowitz and Stegun(1972)}]{besselfn2}
Abramowitz M, Stegun IA (1972) Handbook of Mathematical Functions with
  formulas, graphs, and mathematical tables, 9th edn., Dover Publications, chap
  Section-9.6: Modified Bessel Functions, p 374–377

\bibitem[{Andjamba et~al(2016)Andjamba, Zodi, and Jat}]{namibiainterference}
Andjamba TS, Zodi GAL, Jat DS (2016) Interference analysis of ieee 802.11
  wireless networks: A case study of namibia university of science and
  technology. In: 2016 International Conference on ICT in Business Industry
  Government (ICTBIG), pp 1--5, \doi{10.1109/ICTBIG.2016.7892726}

\bibitem[{Liu et~al(2020)Liu, Aoki, Li, Pei, Choi, Nguyen, and
  Sekiya}]{InterNetworkInterference}
Liu J, Aoki T, Li Z, et~al (2020) Throughput analysis of ieee 802.11 wlans with
  inter-network interference. Applied Sciences 10(6).
  \doi{10.3390/app10062192},
  \urlprefix\url{https://www.mdpi.com/2076-3417/10/6/2192}

\bibitem[{WAN et~al(2016)WAN, SANADA, KOMURO, MOTOYOSHI, YAMAGAKI, SHIODA,
  SAKATA, MURASE, and SEKIYA}]{airtimeconcpet}
WAN Y, SANADA K, KOMURO N, et~al (2016) Throughput analysis of wlans in
  saturation and non-saturation heterogeneous conditions with airtime concept.
  IEICE Transactions on Communications E99.B(11):2289--2296.
  \doi{10.1587/transcom.2016NEP0010}

\bibitem[{Chandaliya et~al(2012)Chandaliya, Dhakate, Lokhande, and
  Ambawade}]{BTInterference}
Chandaliya P, Dhakate N, Lokhande U, et~al (2012) Interference analysis of ieee
  802.11n. In: 2012 International Conference on Communication, Information
  Computing Technology (ICCICT), pp 1--6, \doi{10.1109/ICCICT.2012.6398169}

\bibitem[{Natarajan et~al(2016)Natarajan, Zand, and Nabi}]{multiInterference}
Natarajan R, Zand P, Nabi M (2016) Analysis of coexistence between ieee 802.15.
  4, ble and ieee 802.11 in the 2.4 ghz ism band. In: IECON 2016 - 42nd Annual
  Conference of the IEEE Industrial Electronics Society, 24-27 October 2016,
  Firenze, Italy. Institute of Electrical and Electronics Engineers, United
  States, pp 6025--6032, \doi{10.1109/IECON.2016.7793984},
  \urlprefix\url{http://www.iecon2016.org/?jjj=1483537444731,
  http://www.iecon2016.org/}, 42nd Annual Conference of the IEEE Industrial
  Electronics Society (IECON 2016), IECON 2016 ; Conference date: 24-10-2016
  Through 27-10-2016

\bibitem[{Sarkar et~al(2021)Sarkar, Mussa, and Gul}]{peoplemovement}
Sarkar NI, Mussa O, Gul S (2021) Impact of people’s movement on wi-fi link
  throughput in indoor propagation environments: An empirical study.
  Electronics 10(7):856. \doi{10.3390/electronics10070856}

\bibitem[{Wu et~al(2008)Wu, Lee, Tseng, Jan, and Chuang}]{RSSI}
Wu RH, Lee YH, Tseng HW, et~al (2008) Study of characteristics of rssi signal.
  In: 2008 IEEE International Conference on Industrial Technology, pp 1--3,
  \doi{10.1109/ICIT.2008.4608603}

\bibitem[{Byrne et~al(2018)Byrne, Kozlowski, Santos-Rodriguez, Piechocki, and
  Craddock}]{RSSIlocalization}
Byrne D, Kozlowski M, Santos-Rodriguez R, et~al (2018) Residential wearable
  rssi and accelerometer measurements with detailed location annotations.
  Scientific Data 5(1). \doi{10.1038/sdata.2018.168}

\bibitem[{Rosu(2015)}]{AGC1}
Rosu I (2015) Automatic gain control (agc) in receivers.
  \urlprefix\url{https://www.qsl.net/va3iul/Files/Automatic_Gain_Control.pdf}

\bibitem[{Li et~al(2020)Li, Yang, Yu, Liao, and Xu}]{AGC2}
Li Y, Yang L, Yu L, et~al (2020) Digital agc circuit design based on fpga.
  Journal of Physics: Conference Series 1654(1):012030.
  \doi{10.1088/1742-6596/1654/1/012030}

\bibitem[{Kang and No(2017)}]{AGC3}
Kang H, No JS (2017) Automatic gain control in high adjacent channel
  interference for ofdm systems. In: 2017 23rd Asia-Pacific Conference on
  Communications (APCC), pp 1--4, \doi{10.23919/APCC.2017.8303964}

\bibitem[{Sure and Bhuma(2017)}]{ChE1}
Sure P, Bhuma CM (2017) A survey on ofdm channel estimation techniques based on
  denoising strategies. Engineering Science and Technology, an International
  Journal 20(2):629–636. \doi{10.1016/j.jestch.2016.09.011}

\bibitem[{Kaur et~al(2018)Kaur, Khosla, and Sarin}]{ChE2}
Kaur H, Khosla M, Sarin R (2018) Channel estimation in mimo-ofdm system: A
  review. In: 2018 Second International Conference on Electronics,
  Communication and Aerospace Technology (ICECA), pp 974--980,
  \doi{10.1109/ICECA.2018.8474747}

\bibitem[{Iserte et~al(2005)Iserte, \'{A}ngel Lagunas Hern\'{a}ndez~Miguel, and
  Ana}]{phdthesis}
Iserte AP, \'{A}ngel Lagunas Hern\'{a}ndez~Miguel, Ana IPN (2005) Channel state
  information and joint transmitter-receiver design in multi-antenna systems.
  PhD thesis, Universitat Polit\`{e}cnica de Catalunya

\bibitem[{Gao et~al(2006)Gao, Ozdural, Ardalan, and Liu}]{csi}
Gao J, Ozdural O, Ardalan S, et~al (2006) Performance modeling of mimo ofdm
  systems via channel analysis. IEEE Transactions on Wireless Communications
  5(9):2358--2362. \doi{10.1109/TWC.2006.1687758}

\bibitem[{Pedregosa et~al(2011)Pedregosa, Varoquaux, Gramfort, Michel, Thirion,
  Grisel, Blondel, Prettenhofer, Weiss, Dubourg, Vanderplas, Passos,
  Cournapeau, Brucher, Perrot, and Duchesnay}]{sklearn}
Pedregosa F, Varoquaux G, Gramfort A, et~al (2011) Scikit-learn: Machine
  learning in {P}ython. Journal of Machine Learning Research 12:2825--2830

\bibitem[{CyberZHG(2020)}]{posembd}
CyberZHG (2020) Cyberzhg/keras-pos-embd: Position embedding layers in keras.
  \urlprefix\url{https://github.com/CyberZHG/keras-pos-embd}

\bibitem[{Cho et~al(2014)Cho, van Merrienboer, Bahdanau, and Bengio}]{gru}
Cho K, van Merrienboer B, Bahdanau D, et~al (2014) On the properties of neural
  machine translation: Encoder-decoder approaches. \eprint{1409.1259}

\bibitem[{Shen et~al(2018)Shen, Tan, Zhang, Zeng, and Xu}]{grufin}
Shen G, Tan Q, Zhang H, et~al (2018) Deep learning with gated recurrent unit
  networks for financial sequence predictions. Procedia Computer Science
  131:895–903. \doi{10.1016/j.procs.2018.04.298}

\bibitem[{Schuster and Paliwal(1997)}]{bidirectional}
Schuster M, Paliwal K (1997) Bidirectional recurrent neural networks. IEEE
  Transactions on Signal Processing 45(11):2673--2681. \doi{10.1109/78.650093}

\bibitem[{Vaswani et~al(2017)Vaswani, Shazeer, Parmar, Uszkoreit, Jones, Gomez,
  Kaiser, and Polosukhin}]{attention}
Vaswani A, Shazeer N, Parmar N, et~al (2017) Attention is all you need. In:
  Guyon I, Luxburg UV, Bengio S, et~al (eds) Advances in Neural Information
  Processing Systems, vol~30. Curran Associates, Inc.,
  \urlprefix\url{https://proceedings.neurips.cc/paper/2017/file/3f5ee243547dee91fbd053c1c4a845aa-Paper.pdf}

\bibitem[{Hassan et~al(2022)Hassan, Shams, Hikal, and Elmougy}]{adambest}
Hassan E, Shams MY, Hikal NA, et~al (2022) The effect of choosing optimizer
  algorithms to improve computer vision tasks: A comparative study. Multimedia
  Tools and Applications \doi{10.1007/s11042-022-13820-0}

\bibitem[{Kingma and Ba(2014)}]{adamoriginal}
Kingma DP, Ba J (2014) Adam: A method for stochastic optimization.
  \doi{10.48550/ARXIV.1412.6980},
  \urlprefix\url{https://arxiv.org/abs/1412.6980}

\bibitem[{Ajagekar(2021)}]{adameasy}
Ajagekar A (2021) Adam optimizer.
  \urlprefix\url{https://optimization.cbe.cornell.edu/index.php?title=Adam}

\bibitem[{Ben-David et~al(2009)Ben-David, Blitzer, Crammer, Kulesza, Pereira,
  and Vaughan}]{domain_shift}
Ben-David S, Blitzer J, Crammer K, et~al (2009) A theory of learning from
  different domains. Machine Learning 79(1-2):151–175.
  \doi{10.1007/s10994-009-5152-4}

\bibitem[{Habrard et~al(2019)Habrard, Youn\`{e}s, Morvant, Redko, and
  Sebban}]{domain_adaptation1}
Habrard A, Youn\`{e}s B, Morvant E, et~al (2019) Advances in domain adaptation
  theory, 1st edn., Elsevier, chap 1-9, pp 1--208

\bibitem[{Redko et~al(2020)Redko, Morvant, Habrard, Sebban, and
  Bennani}]{domain_adaptation2}
Redko I, Morvant E, Habrard A, et~al (2020) A survey on domain adaptation
  theory: learning bounds and theoretical guarantees. \eprint{2004.11829}

\bibitem[{Xie et~al(2015)Xie, Li, and Li}]{csi_tool_atheros}
Xie Y, Li Z, Li M (2015) Precise power delay profiling with commodity wifi.
  ACM, pp 53--64, \doi{10.1145/2789168.2790124}

\bibitem[{Kaveh and Mesgari(2022)}]{MHopt1}
Kaveh M, Mesgari MS (2022) Application of meta-heuristic algorithms for
  training neural networks and deep learning architectures: A comprehensive
  review. Neural Processing Letters \doi{10.1007/s11063-022-11055-6}

\bibitem[{Abd~Elaziz et~al(2021)Abd~Elaziz, Dahou, Abualigah, Yu, Alshinwan,
  Khasawneh, and Lu}]{MHopt2}
Abd~Elaziz M, Dahou A, Abualigah L, et~al (2021) Advanced metaheuristic
  optimization techniques in applications of deep neural networks: A review.
  Neural Computing and Applications 33(21):14079–14099.
  \doi{10.1007/s00521-021-05960-5}

\bibitem[{Khan et~al(2021)Khan, Jabeen, Ghouzali, Rehman, Naz, and
  Abdul}]{MHopt3}
Khan MS, Jabeen F, Ghouzali S, et~al (2021) Metaheuristic algorithms in
  optimizing deep neural network model for software effort estimation. IEEE
  Access 9:60309–60327. \doi{10.1109/access.2021.3072380}

\bibitem[{Movassagh et~al(2021)Movassagh, Alzubi, Gheisari, Rahimi, Mohan,
  Abbasi, and Nabipour}]{MHopt4}
Movassagh AA, Alzubi JA, Gheisari M, et~al (2021) Artificial neural networks
  training algorithm integrating invasive weed optimization with differential
  evolutionary model. Journal of Ambient Intelligence and Humanized Computing
  \doi{10.1007/s12652-020-02623-6}

\bibitem[{Xu et~al(2020)Xu, Panwar, Kodialam, and Lakshman}]{DP1}
Xu S, Panwar SS, Kodialam M, et~al (2020) Deep neural network approximated
  dynamic programming for combinatorial optimization. Proceedings of the AAAI
  Conference on Artificial Intelligence 34(02):1684–1691.
  \doi{10.1609/aaai.v34i02.5531}

\bibitem[{Wu and Wang(2018)}]{DP2}
Wu N, Wang H (2018) Deep learning adaptive dynamic programming for real time
  energy management and control strategy of micro-grid. Journal of Cleaner
  Production 204:1169–1177. \doi{10.1016/j.jclepro.2018.09.052}

\bibitem[{Alzubi et~al(2020)Alzubi, Alzubi, Alweshah, Qiqieh, Al-Shami, and
  Ramachandran}]{DP3}
Alzubi OA, Alzubi JA, Alweshah M, et~al (2020) An optimal pruning algorithm of
  classifier ensembles: Dynamic programming approach. Neural Computing and
  Applications 32(20):16091–16107. \doi{10.1007/s00521-020-04761-6}

\bibitem[{Gheisari et~al(2021)Gheisari, Najafabadi, Alzubi, Gao, Wang, Abbasi,
  and Castiglione}]{obpp}
Gheisari M, Najafabadi HE, Alzubi JA, et~al (2021) Obpp: An ontology-based
  framework for privacy-preserving in iot-based smart city. Future Generation
  Computer Systems 123:1–13. \doi{10.1016/j.future.2021.01.028}

\bibitem[{Hanzo et~al(2011)Hanzo, Akhtman, and Wang}]{hanzo}
Hanzo L, Akhtman YJ, Wang L (2011) MIMO-OFDM for LTE, Wi-Fi and WiMAX: Coherent
  versus non-coherent and cooperative turbo-transceivers, Wiley, chap
  Chapter-1: Introduction to OFDM and MIMO-OFDM, Section-7.8: Channel
  Estimation for MIMO-OFDM, pp 1, 233–33, 244

\end{thebibliography}

\end{document}